\documentclass[11pt, a4paper, logo]{template}
\usepackage{fancyhdr}

\usepackage{natbib}
\usepackage{subcaption,booktabs}
\usepackage{tabularx}
\usepackage{xspace}
\usepackage{amsmath}
\usepackage{graphicx}
\usepackage{caption}
\usepackage{cleveref}
\usepackage{url}
\usepackage{framed} 
\usepackage{multirow}
\usepackage{booktabs}
\usepackage[export]{adjustbox}
\usepackage{wrapfig}
\usepackage{tablefootnote}
\usepackage{floatrow}
\usepackage{makecell}
\usepackage{multirow,booktabs}
\usepackage{amsmath,amssymb} 
\usepackage{pifont}
\usepackage{color}
\usepackage{xurl}
\usepackage{url}
\usepackage{longtable}
\usepackage{natbib}
\usepackage{bbm}
\usepackage{abstract}
\setcitestyle{square,numbers}
\newfloatcommand{capbtabbox}{table}[][\FBwidth]
\newfloatcommand{capbfigbox}{figure}[][\FBwidth]

\newcommand{\cmark}{\ding{51}}%
\newcommand{\xmark}{\ding{55}}%

\newcommand{\etc}{\textit{etc.}}%
\newcommand{\eg}{\textit{e.g.}}%
\definecolor{baselinecolor}{gray}{.9}

\newcommand{\underfigtab}{\vspace{0pt}}

\usepackage[justification=raggedright,font=small]{caption}
\title{Open-sourced Data Ecosystem in Autonomous Driving: The Present and Future}

\author[ \hspace{-0.6ex}]{
{OpenDriveLab$^1$} \quad 
Shanghai AI Lab$^2$ \quad
\mbox{Star League}$^3$
}
\affil[]{
Presented by multi-lateral collaborations from industry and academia institutes.
Project led by OpenDriveLab. $^*$Equal contributions. 
$^\dagger$Primary contact: 
\texttt{hy@opendrivelab.com},
\texttt{qiaoyu@pjlab.org.cn}
}
\chineseversion{This article is an English translation of the Survey written in Chinese. 
In case of ambiguity, the 
\href{https://opendrivelab.com/Dataset_Survey_Chinese_V2.pdf}{Chinese version}
shall prevail.
\\ \smallskip 
$^2$ OpenDriveLab is part of Shanghai AI Lab.\\
$^3$ {Star League} is a cohort of global entities represented by a wide diversity to advance the development of Autonomous driving, Robotics, Embodied AI and beyond. For more details, see \url{https://starleague.ai}.
}

\paperurl{https://arxiv.org/abs/2312.03408}

\renewenvironment{abstract}
{\par\noindent\\ \ignorespaces \bfseries}
{\par\bigskip}

\begin{document}

\maketitle

\vspace{-25pt}
\noindent 

\begin{table}[h]
    \begin{adjustwidth}{-0.03\textwidth}{}
      \centering
      \renewcommand{\arraystretch}{1.35}
      \resizebox{.9\textwidth}{!}{
       
          \begin{tabular}{clclclcl}
\multirow{2}{*}{\includegraphics[height=1.1cm, valign=c]{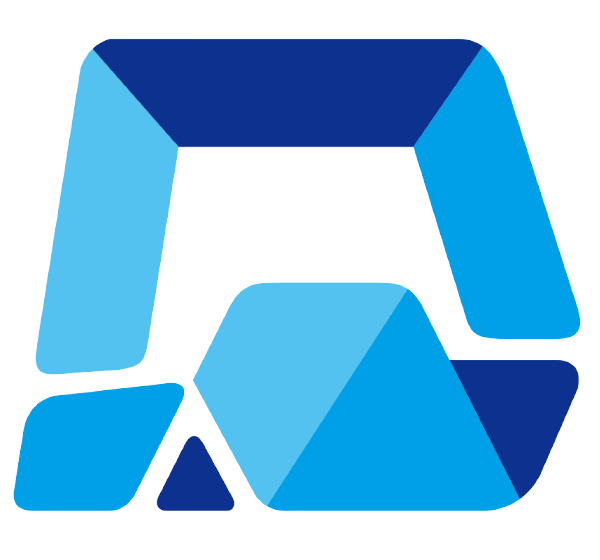}}  & \textbf{Hongyang Li} $^{1,2,3~* \dagger}$ 
& \multirow{2}{*}{\includegraphics[height=1.1cm, valign=c]{figures/affiliation/opendrivelab.png}} & \textbf{Yang Li} $^{1,2~*}$           
& \multirow{2}{*}{\includegraphics[height=1.1cm, valign=c]{figures/affiliation/opendrivelab.png}} & \textbf{Huijie Wang}  $^{1,2~*}$    
& \multirow{2}{*}{\includegraphics[height=1.1cm, valign=c]{figures/affiliation/opendrivelab.png}} & \textbf{Jia Zeng}  $^{1,2~*}$                  \\
& OpenDriveLab   &      & OpenDriveLab   &     & OpenDriveLab    &    & OpenDriveLab      

\\
\\

\multirow{2}{*}{\includegraphics[height=1.1cm, valign=c]{figures/affiliation/opendrivelab.png}}    & \textbf{Huilin Xu} $^{1,2}$       
& \multirow{2}{*}{\includegraphics[height=1.cm, valign=c]{figures/affiliation/opendrivelab.png}} & \textbf{Pinlong Cai} $^{1,2}$      
& \multirow{2}{*}{\includegraphics[height=1.1cm, valign=c]{figures/affiliation/opendrivelab.png}} & \textbf{Li Chen} $^{1,2,3}$          
& \multirow{2}{*}{\includegraphics[height=1.4cm, valign=c]{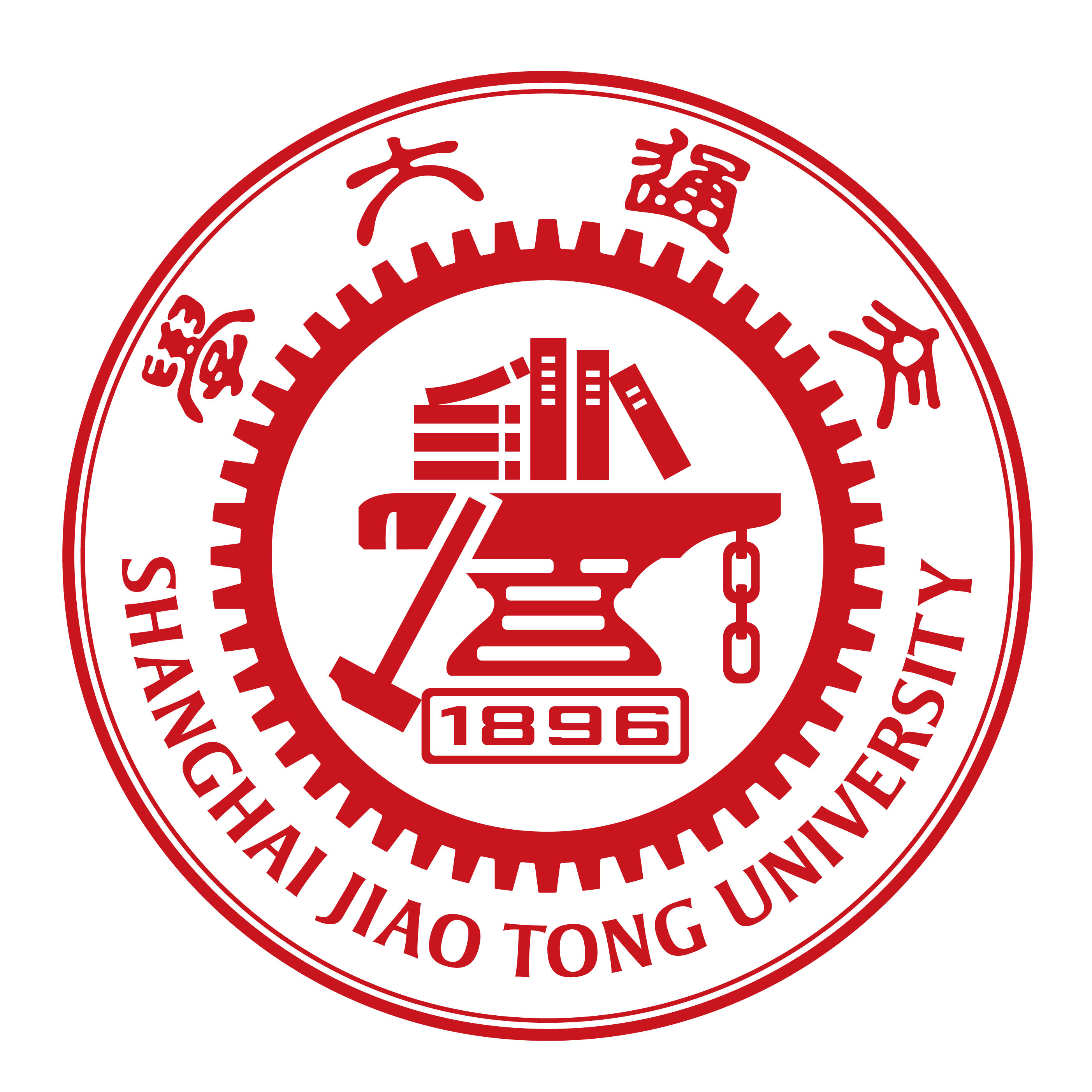}} & \textbf{Junchi Yan} $^{3}$                   \\
& Shanghai AI Lab   &     & Shanghai AI Lab  &    & Shanghai AI Lab   &    & SJTU  
\\
\\
\multirow{2}{*}{\includegraphics[height=1.1cm, valign=c]{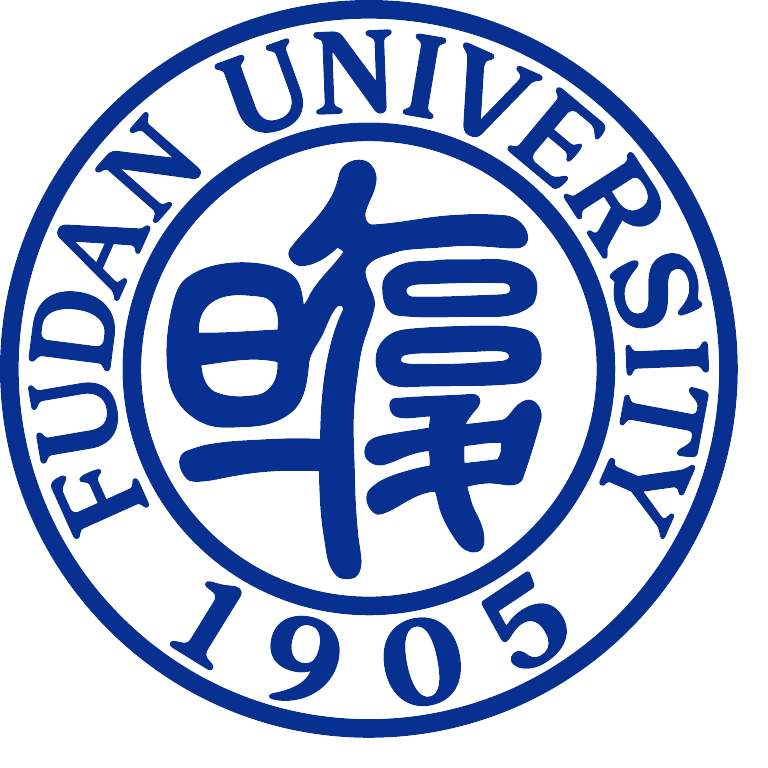}} & \textbf{Feng Xu} $^{3}$            
& \multirow{2}{*}{\includegraphics[height=1.2cm, valign=c]{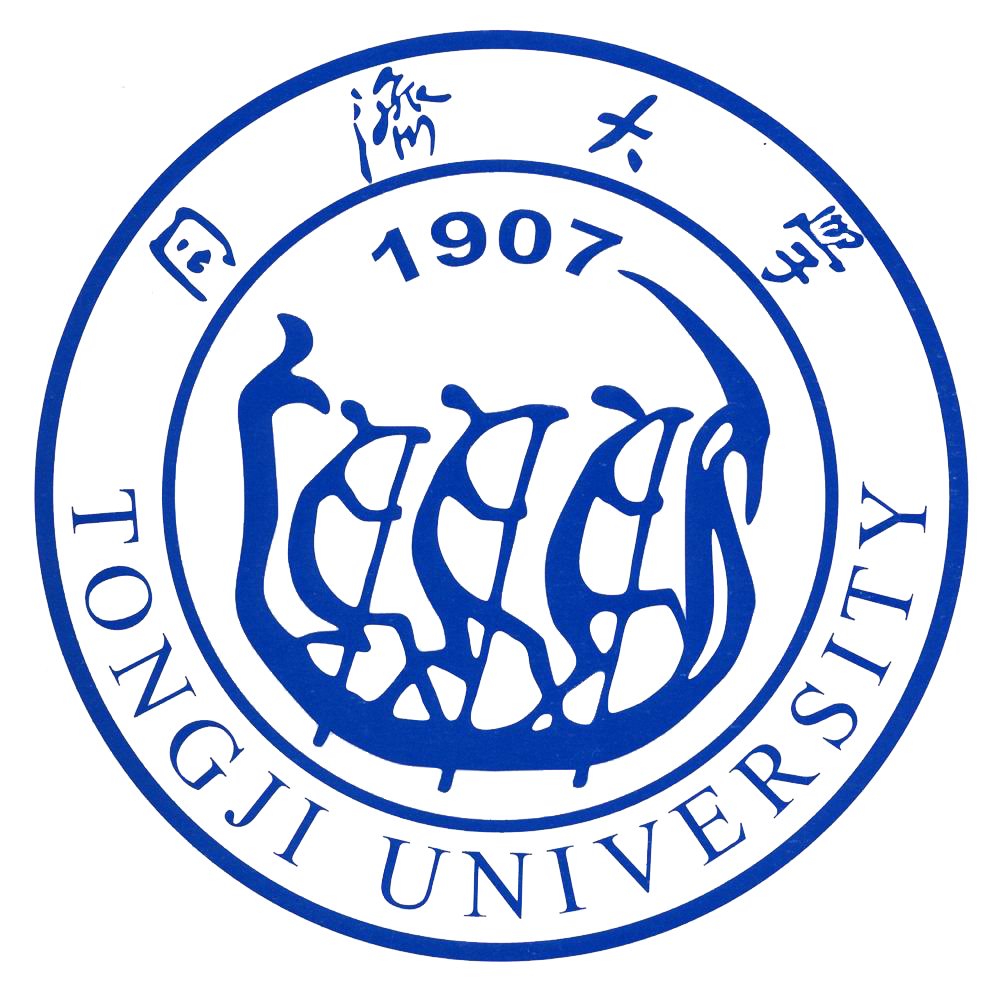}} & \textbf{Lu Xiong} $^{3}$         
& \multirow{2}{*}{\includegraphics[height=1.1cm, valign=c]{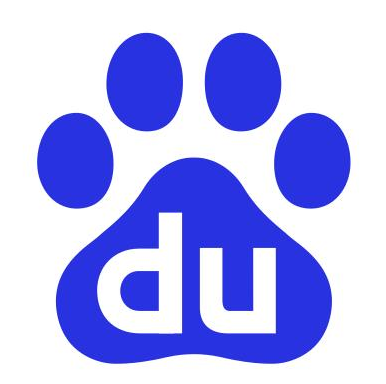}} & \textbf{Jingdong Wang} $^{3}$    
& \multirow{2}{*}{\includegraphics[height=1.1cm, valign=c]{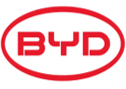}} & \textbf{Futang Zhu} $^{3}$                   
\\
& Fudan University   &    & Tongji University &    & Baidu     &     
& BYD Auto               
\\
\\
\multirow{2}{*}{\includegraphics[height=1.3cm, valign=c]{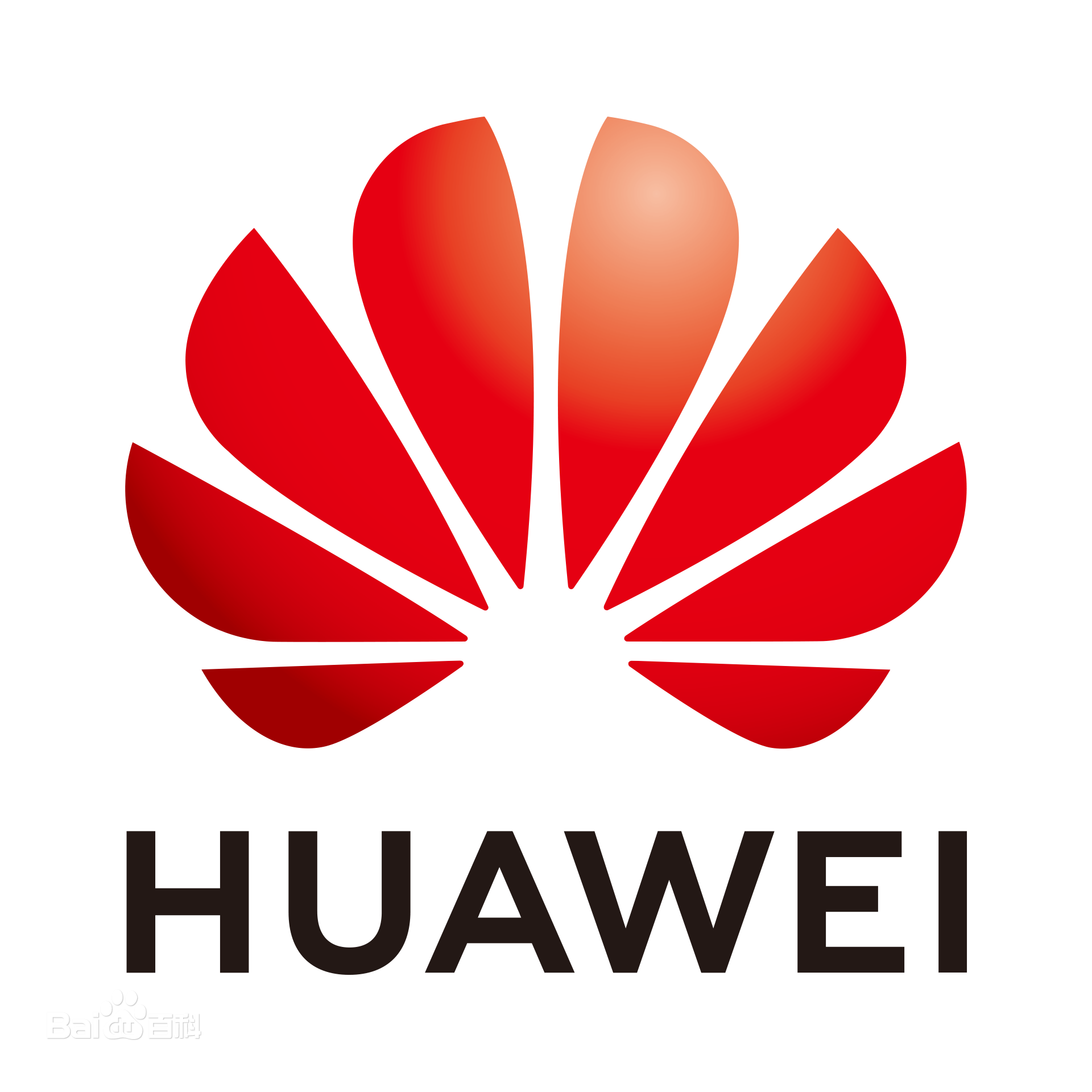}} & \textbf{Chunjing Xu} $^{3}$      
& \multirow{2}{*}{\includegraphics[height=0.9cm, valign=c]{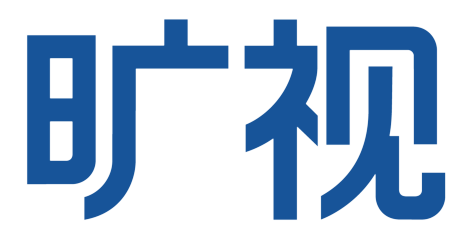}} &\textbf{Tiancai Wang} $^{3}$    
& \multirow{2}{*}{\includegraphics[height=1.1cm, valign=c]{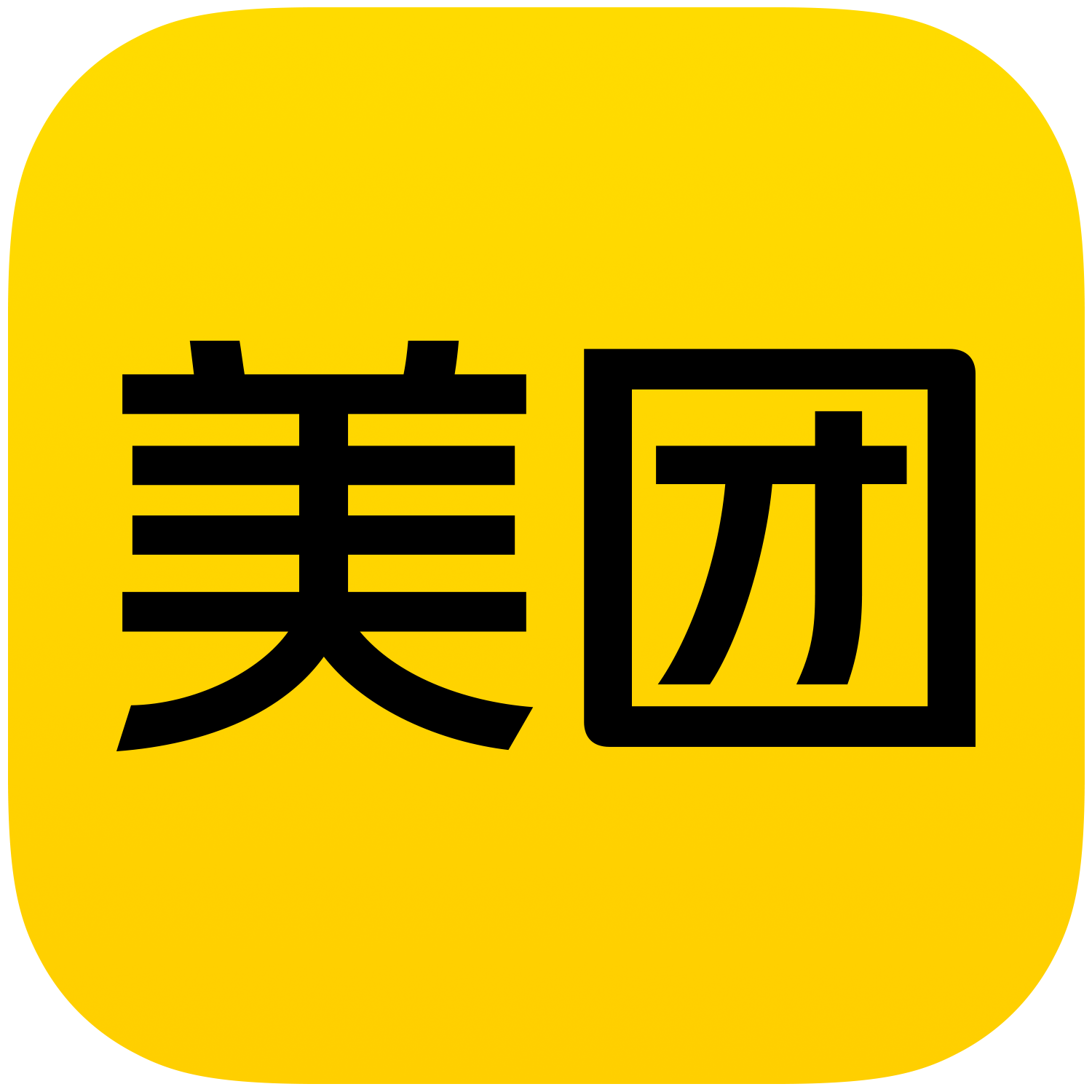}} & \textbf{Fei Xia}  $^{3}$ 
&  \multirow{2}{*}{\includegraphics[height=1.1cm, valign=c]{figures/affiliation/meituan.png}} & \textbf{Beipeng Mu}   $^{3}$ 
\\
   & Huawei         &     & MEGVII &   & Meituan & & Meitua   
\\
\\

\multirow{2}{*}{\includegraphics[height=1.2cm, valign=c]{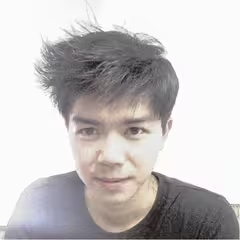}} & \textbf{Zhihui Peng}  $^{3}$    
& \multirow{2}{*}{\includegraphics[height=1.cm, valign=c]{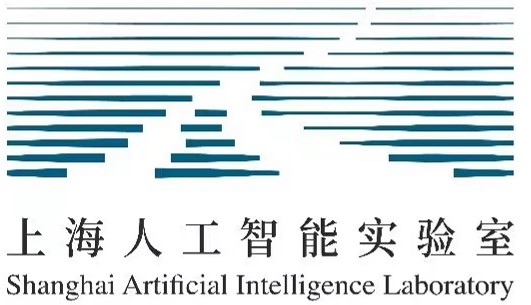}} & \textbf{Dahua Lin}  $^{2,3}$       
& \multirow{2}{*}{\includegraphics[height=1.cm, valign=c]{figures/affiliation/ailab.jpg}} &\textbf{Yu Qiao}    $^{2,3~\dagger}$                    
\\
 & AGIBOT      &    & Shanghai AI Lab    &    & Shanghai AI Lab
\end{tabular}        
      }
    \end{adjustwidth}
\end{table}

\begin{abstract}
With the growing 
development
of autonomous driving, a systematic examination of open-source autonomous driving datasets becomes instrumental in fostering the robust evolution of the industry ecosystem. Current 
datasets can be broadly categorized into two generations. The first-generation 
features
approximately
simpler sensor modalities, smaller data scale, and is limited to perception-level tasks. KITTI, introduced in 2012, serves as a prominent representative of this initial wave.
In contrast, the second-generation 
exhibits heightened complexity in sensor modalities, greater data scale and diversity, and an expansion of tasks from perception to encompass prediction and control. Leading examples of the second generation include nuScenes and Waymo proposed around 2019.
In this work,
we 
assess over seventy open-source autonomous driving datasets from 
a wide span of 
sources. It offers insights into various aspects, such as the principles underlying the creation of high-quality datasets, the pivotal role of data engine systems, and the utilization of generative foundation models to facilitate scalable data generation.
This review undertakes a further analysis and discourse regarding the key essentials and data scale that the forthcoming third-generation 
datasets should possess. It delves into the scientific and technical challenges that warrant resolution as well.  
%
For more details, please refer to \url{https://github.com/OpenDriveLab/DriveAGI}.
\end{abstract}


\section{Introduction} \label{sec:intro}

As the trending advancements of artificial intelligence (AI) unfold, a new wave of international competition is triggered among  nations that regard AI as a national strategy~\citep{Euro2020ai, National2023ai, council2017ai}.
As one of the important areas of AI, autonomous driving is expected to reshape the existing traffic and transport patterns, which will greatly improve efficiency and safety and have a profound impact on the future development of cities and society.

Autonomous driving requires a large amount of data for model training to perceive and understand the traffic environment in order to make correct decisions and actions. 
\textbf{The construction of autonomous driving datasets is crucial for the development of autonomous driving technology.}
For instance, in the second quarter of 2023, Tesla, an autonomous driving company in the United States, tested its Full Self-Driving Beta with an unprecedented total driving mile of 300 million miles (about 483 million kilometers), which will continuously grow as the number of users increases. The massive amount of data collected from users is the key factor for Tesla to maintain its advancements in the area of autonomous driving.
On the other side, \textbf{the emergence of foundation models in natural language processing and generalized vision, which demonstrates the importance of massive high-quality data, also inspires us with the necessity for constructing autonomous driving datasets.}

\begin{figure}[t]
    \centering
    \includegraphics[width=.9\linewidth]{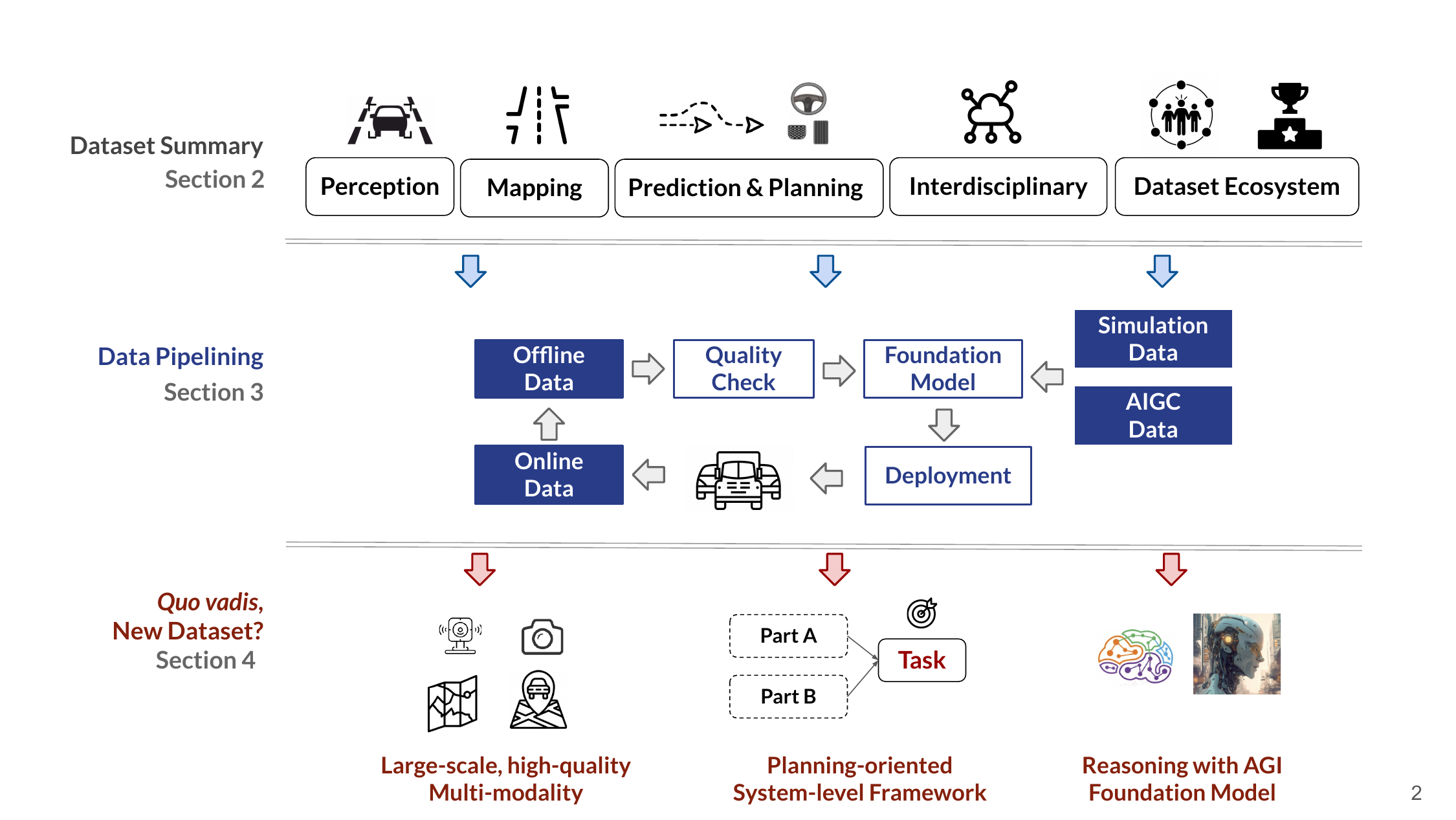}
    \caption{
        \textbf{Survey at a glance.}
        \Cref{sec:Autonomous_Driving_Datasets} summarizes details of approximately 70 existing autonomous driving datasets in the wild. 
        \Cref{sec:Data_Engine} discusses the data engine system solution. 
        \Cref{sec:Next-generation_Autonomous_Driving_Dataset} delineates key essentials for building the next-generation dataset. 
    }
    \label{fig:Layout_survey}
    \underfigtab
\end{figure}

\Cref{fig:Layout_survey} depicts the main content and structure of this survey. 
\Cref{sec:Autonomous_Driving_Datasets} divides autonomous driving datasets into different categories, namely perception, mapping, and planning datasets. We further discuss the current status and development of each type of dataset based on impact, community ecosystem, and international challenges.
In \Cref{sec:Data_Engine}, we describe data engine systems from various companies, with a focus on data labeling, quality control, simulation, and data generation and auto-labeling based on foundation models. \Cref{sec:Next_Generation} concludes and provides a vision of the next-generation autonomous driving datasets.

\begin{figure}[t]
    \centering
    \includegraphics[width=1.0\linewidth]{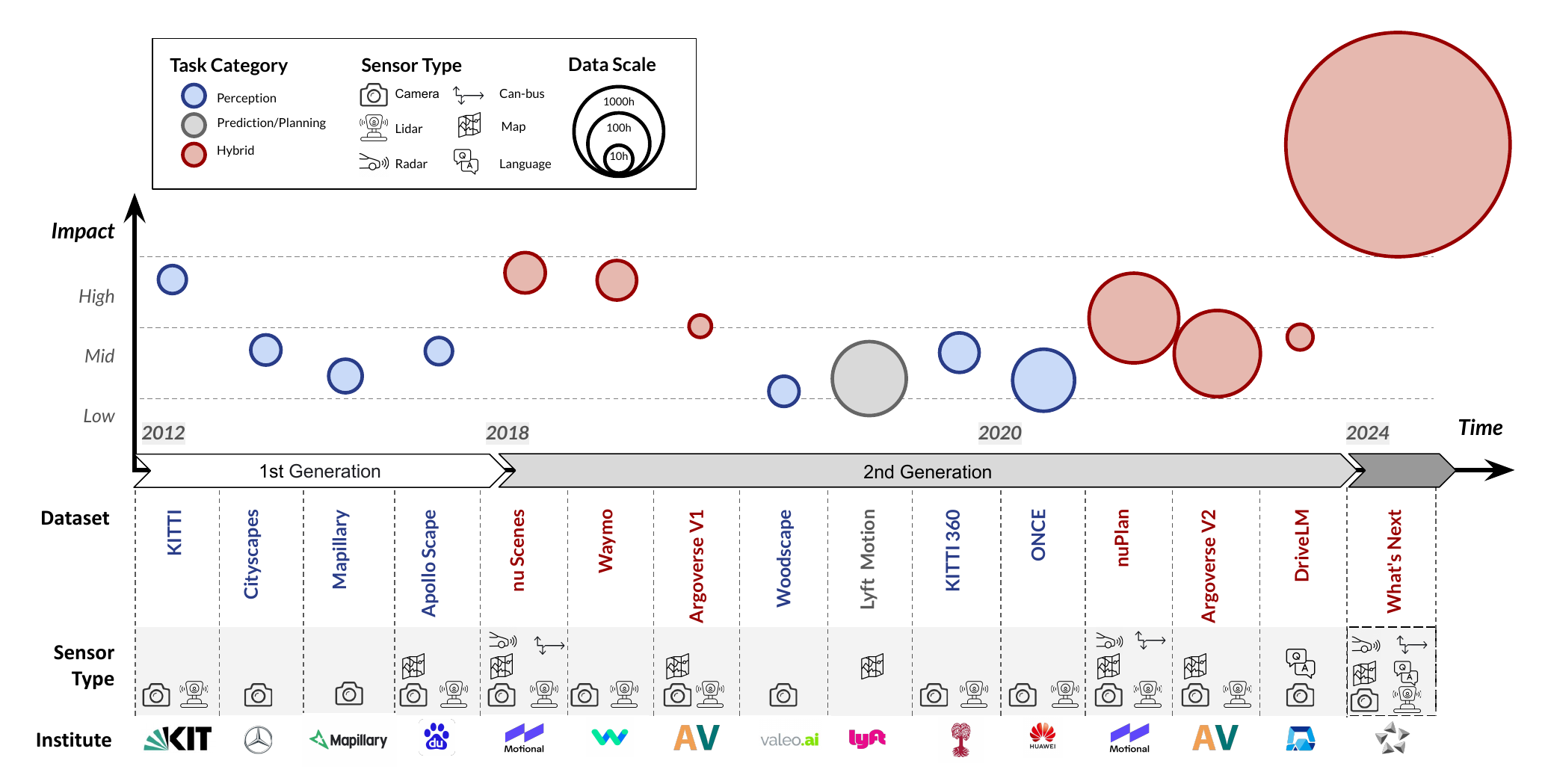}
    \caption{
    \textbf{Roadmap of Autonomous Driving Datasets.}
        We present popular datasets since the 2010s chronologically, and categorize them in terms of Impact. 
        The Impact of a dataset is defined based on sensor setup, input modality, task, data scale, ecosystem, \etc
    }
    \label{fig:Roadmap_datasets}
    \underfigtab
\end{figure}

In this paper, we provide a systematic survey on existing driving autonomous datasets.
\Cref{fig:Roadmap_datasets} illustrates the development of public datasets chronologically. 
Based on the comprehensive factors, including types of sensors, data scale, scene diversity, and supported tasks, we classify current datasets into two generations: 
representing the first generation of datasets, KITTI~\citep{Dataset_kitti}, which was proposed in 2012 with a front-view camera and a LiDAR, supported a series of tasks that bring autonomous driving into public vision; 
the second generation of datasets, represented by the nuScenes~\citep{Dataset_nuScenes} and Waymo~\citep{Dataset_Waymo} dataset, comprises broader types of sensors, such as multi-view cameras, radar, localization information, and high-definition maps, that enable a larger variety of downstream tasks, \eg, perception, mapping, prediction, and planning.
With the development of autonomous driving technology, open-source datasets contain more types of sensors, larger scale of data, and more diverse scenes, and thus enable more complex tasks of autonomous driving.

In the era of large language models (LLMs) and vision foundation models, new challenges, as well as chances, emerge when establishing new autonomous driving datasets. 
Traditionally, constructing a dataset requires multiple steps, including sensor calibration, data collection, annotation, and data cleaning.
This scheme suffers from high costs for data collection and human annotation, and it is difficult to conduct quality control. 
Thus, researchers begin to pay attention to the method for building a new dataset with a limited cost but higher quality.
The appearance of Artificial Intelligence Generated Content (AIGC) provides a new possibility.
AIGC is able to generate virtual traffic environments with multiple types of traffic agents, various road signs and lanelines, and abundant weather conditions.
By utilizing this advantage, rare data, such as in uncommon weather conditions and long-tail scenarios, can be acquired to reduce cost and accelerate the research process.
However, as the traffic scenarios generated by AIGC still have a domain gap compared to realistic data, the data quality and applicability of AIGC technology need to be discussed.

As autonomous driving technology progressively evolves and the complexity of tasks increases, it is necessary to establish the next-generation datasets, which comprise data with better quality and larger scale.
By summarizing current datasets and techniques in \Cref{sec:Autonomous_Driving_Datasets} and \Cref{sec:Data_Engine}, we deem that the following properties need to be fulfilled in new autonomous driving datasets: 
(1) full coverage of sensor categories and amounts, abundant data that cover all scenarios, and high-quality raw data and annotation;
(2) flexibility in formulation for the usage in the short term and in the long term, and support for new paradigms, such as end-to-end-frameworks and world models;
(3) intelligence-oriented, which enables verification in interpretability and supports logical reasoning in language.
To sum up, our contributions are as follows:
\begin{enumerate}
    \setlength{\itemsep}{5pt}
    \setlength{\parsep}{0pt}
    \setlength{\parskip}{0pt}
    \item We conduct a comprehensive analysis of existing autonomous driving datasets and evaluate their impact based on a proposed metric.
    \item We analyze the core factors for building high-quality autonomous driving datasets.
    \item Collaborated with the industry, we analyze the practical demands of autonomous driving, and present vision and plan for the next-generation autonomous driving datasets in the era of foundation models.
\end{enumerate}

\section{Autonomous Driving Datasets}\label{sec:Autonomous_Driving_Datasets}

This article analyses close to one hundred datasets from 2012 to the present. \Cref{fig:Roadmap_datasets} presents 14 datasets with significant impact, depicting publication dates on the horizontal axis and impact values on the vertical axis. It provides a qualitative overview of major publicly available datasets. The Y-axis shows the impact values of the datasets, determined according to the criteria defined in \Cref{sec:Impact_of_Datasets_Influences}, taking into account data quality and community ecosystem. From \Cref{tab:dataset_comparison}, it can be seen that KITTI, nuScenes, and Waymo are ranked highest in terms of impact values. Generally, sensor modalities are becoming more complex, data scale and diversity are on the rise, and datasets tasks are expanding from perception to prediction and planning.

\subsection{Impact of Datasets}\label{sec:Impact_of_Datasets_Influences}
We propose an evaluation metric to estimate the impact of autonomous driving datasets on the effectiveness of algorithms' development.The metric aims to evaluate usability, accuracy, and applicability, and there is a current absence of such metrics in the literature despite the abundance of datasets available. To aid researchers and practitioners in selecting datasets relevant to their research queries, this paper presents the datasets impact evaluation metric. A dataset's impact metric comprises of a data quality score $Q_s$ and an community score $C_s$, as depicted in \Cref{eqn:Impact}. The definitions for $Q_s$ and $C_s$ are supplied in \Cref{eqn:Quality_score} and \Cref{eqn:Community_score}, correspondingly.

\begin{equation}
Impact=Q_s+C_s\label{eqn:Impact}
\end{equation}
\begin{equation}
Q_s=\frac{NOC}{5}\cdot\frac{NST-NST_{min}}{NST_{max}-NST_{min}}\cdot\frac{\lg{DD}}{\lg{DD_{max}}}\cdot\frac{\lg{NOS}}{\lg{NOS_{max}}} \label{eqn:Quality_score}
\end{equation}

\begin{equation}
C_s=\mathbbm{1}(NOCH)+\mathbbm{1}(NOLB) \label{eqn:Community_score}
\end{equation}

In \Cref{eqn:Quality_score}, NOC (Number of Continents) indicates the number of continents in which the dataset is distributed, while NST (Number of Sensor Types) represents the quantity of sensor types utilised, such as cameras, LiDAR, and radar. DD(Duration of Dataset) specifies the total duration of the driving dataset in hours, and NOS (Number of Scenes) is the number of scenes. In \Cref{eqn:Community_score}, NOCH (Number of Challenges) denotes the count of related competitions conducted, while NOLB (Number of Leaderboards) represents the tally of task leaderboards maintained. The function $\mathbbm{1}(\cdot)$ returns 1 if the value is greater than or equal to 1 and 0 if it is less than 1.

By evaluating the impact, we classify the presently accessible public datasets into three categories: low, middle, and high. The score range for the low level is [0, 1], for the middle level is [1, 2], and for the high level is [2, 4]. Further details regarding the impact can be found in \Cref{tab:dataset_comparison}.

\subsection{Challenges and Benchmarks}\label{sec:Challenges_and_Benchmarks_in_Autonomous_Driving}

\begin{table}[hp!]
    \renewcommand{\arraystretch}{0.93}
    \caption{
        \textbf{Challenges and Benchmarks in Autonomous Driving.} \texttt{\# Entry} denotes the number of public entries on the leaderboard by the date of manuscript submission (September 2023). 
        \texttt{Test Server} indicates whether the test server remains open or not.
    }
    \centering
    \label{tab:tasks}
    \resizebox{\columnwidth}{!}{
        \footnotesize
        \begin{tabular}{llllccc}
        
            \toprule
            
            \multirow{2}{*}{Title} & \multirow{2}{*}{Host}  & \multirow{2}{*}{Year} & \multirow{2}{*}{Task} & \multirow{2}{*}{Evaluation}   & \multirow{2}{*}{\makecell[c]{\# Entry}} &  \multirow{2}{*}{\makecell[c]{Test\\Server}}\\
            
            &   &  &  &  &     &  \\
             
            \midrule 
            
            \multirow{5}{*}{\makecell[l]{\tiny Autonomous\\\tiny Driving\\\tiny Challenge \\\tiny \citep{adchallenge2023opendrivelab}}}   & \multirow{5}{*}{\tiny OpenDriveLab} & \multirow{5}{*}{\makecell[l]{\tiny CVPR\\\tiny 2023}} 
            &  \textbf{Perception} / OpenLane Topology &OpenLane-V2 Score (OLS)& \multirow{5}{*}{111} &   \cmark \\
            \cmidrule{4-5}
             &   &  
             &  \textbf{Perception} / Online HD Map Construction & mAP &   &    \xmark \\
             \cmidrule{4-5}
            &   & 
            &  \textbf{Perception} / 3D Occupancy Prediction&  mIoU&  &      \cmark \\
            \cmidrule{4-5}
            &   &   
            & \textbf{Prediction \& Planning} / nuPlan Planning& Mean Overall Score &&       \xmark \\

            \arrayrulecolor[gray]{0.9}\specialrule{3pt}{0pt}{-6pt} \arrayrulecolor{black}
            \midrule 
            
        \rowcolor[gray]{0.9}
                       &   &  
                    &  \textbf{Perception} / 2D Video Panoptic Segmentation& weighted Segmentation and Tracking Quality &   & \cmark \\
                    \arrayrulecolor[gray]{0.9}\specialrule{6pt}{0pt}{-6pt} \arrayrulecolor{black}\cmidrule{4-5}
        \rowcolor[gray]{0.9}
                     &   & 
                     &   \textbf{Perception} / Pose Estimation &Pose Estimation Metric (PEM)&  &     \cmark \\
                     \arrayrulecolor[gray]{0.9}\specialrule{6pt}{0pt}{-6pt} \arrayrulecolor{black}\cmidrule{4-5}
        \rowcolor[gray]{0.9}
                    &   & 
                    &  \textbf{Prediction} / Motion Prediction & Soft mAP & &    \cmark \\
                    \arrayrulecolor[gray]{0.9}\specialrule{6pt}{0pt}{-6pt} \arrayrulecolor{black}\cmidrule{4-5}
        \rowcolor[gray]{0.9}
                    &   &   
                    \multirow{-5.5}{*}{\makecell[l]{\tiny CVPR\\\tiny 2023}} & \textbf{Prediction} / Sim Agents  & Realism Meta Metric &  \multirow{-5.5}{*}{35} &   \cmark \\
                    \arrayrulecolor[gray]{0.9}\specialrule{6pt}{0pt}{-6pt} \arrayrulecolor{black}\cmidrule{3-7}
                    
        \rowcolor[gray]{0.9}
                     &  &
                     &  \textbf{Prediction} / Motion Prediction & Soft mAP &   & \cmark \\
                    \arrayrulecolor[gray]{0.9}\specialrule{6pt}{0pt}{-6pt} \arrayrulecolor{black}\cmidrule{4-5}
        \rowcolor[gray]{0.9}
                     &   &  
                     & \textbf{Prediction} / Occupancy and Flow Prediction & AUC on Joint Occupancy and Flow Metric & &  \cmark \\
                     \arrayrulecolor[gray]{0.9}\specialrule{6pt}{0pt}{-6pt} \arrayrulecolor{black}\cmidrule{4-5}
        \rowcolor[gray]{0.9}
                    &   & 
                    &  \textbf{Perception} / 3D Semantic Segmentation &  mIOU &     & \cmark \\
                    \arrayrulecolor[gray]{0.9}\specialrule{6pt}{0pt}{-6pt} \arrayrulecolor{black}\cmidrule{4-5}
        \rowcolor[gray]{0.9}
                    &   &   
                     \multirow{-5.5}{*}{\makecell[l]{\tiny CVPR\\\tiny 2022}} &  \textbf{Perception} / 3D Camera-only Detection & Longitudinal Affinity Weighted LET-3D-AP   &  \multirow{-5.5}{*}{128} & \cmark \\
                    
                    \arrayrulecolor[gray]{0.9}\specialrule{6pt}{0pt}{-6pt} \arrayrulecolor{black}\cmidrule{3-7}
        \rowcolor[gray]{0.9}
                     &  &  
                     &  \textbf{Prediction} / Motion Prediction& Soft mAP&   & \cmark \\
                    \arrayrulecolor[gray]{0.9}\specialrule{6pt}{0pt}{-6pt} \arrayrulecolor{black}\cmidrule{4-5}
        \rowcolor[gray]{0.9}
                     &   &  
                     & \textbf{Predirction} / Interaction Prediction & mAP &  & \cmark \\
                     \arrayrulecolor[gray]{0.9}\specialrule{6pt}{0pt}{-6pt} \arrayrulecolor{black}\cmidrule{4-5}
        \rowcolor[gray]{0.9}
                    &   &  
                    &  \textbf{Perception} / Real-time 3D Detection  & Mean Average Precision with Heading (APH)&      & \cmark \\
                    \arrayrulecolor[gray]{0.9}\specialrule{6pt}{0pt}{-6pt} \arrayrulecolor{black}\cmidrule{4-5}
        \rowcolor[gray]{0.9}
                    \multirow{-18}{*}{\makecell[l]{\tiny Waymo\\\tiny Open\\\tiny Dataset\\\tiny Challenges\\\tiny \citep{ Dataset_Waymo, adchallenge2023Waymo}}} &  \multirow{-18}{*}{\tiny Waymo} & \multirow{-5.5}{*}{\makecell[l]{\tiny CVPR\\\tiny 2021}}  
                    &  \textbf{Perception} / Real-time 2D Detection &  mAP & \multirow{-5.5}{*}{115}  & \cmark \\
\arrayrulecolor[gray]{0.9}\specialrule{3pt}{0pt}{-3pt} \arrayrulecolor{black}
                    \midrule 
                    \multirow{18}{*}{\makecell[l]{\tiny Argoverse\\\tiny Challenges\\\tiny \citep{Dataset_argoversev2,Dataset_argoversev1}}}    & \multirow{18}{*}{\tiny Argoverse}  & \multirow{8}{*}{\makecell[l]{\tiny CVPR\\\tiny 2023}} 
                    & \textbf{Prediction} / Multi-agent Forecasting  & \makecell[c]{Average Brier Minimum Final Displacement\\Error (avgBrierMinFDE)}&\multirow{8}{*}{81}  & \cmark \\
                    \cmidrule{4-5}
                     &   & 
                     &  \makecell[l]{\textbf{Perception \& Prediction} / Unified Sensor-\\based Detection, Tracking, and Forecasting } & Forecasting Average Precision &    & \cmark \\
                     \cmidrule{4-5}
                     &   &  
                     & \textbf{Perception} / LiDAR Scene Flow& \makecell[c]{Three Way Average End Point Error} &&     \cmark \\
                     \cmidrule{4-5}
                     &   &  
                     & \textbf{Prediction} / 3D Occupancy Forecasting & \makecell[c]{L1 Error (L1), Absolute Relative L1 Error\\(AbsRel), Near-field Chamfer Distance (NFCD)}&    & \cmark \\
        
                    \cmidrule{3-7}
        
                     &  & \multirow{5}{*}{\makecell[l]{\tiny CVPR\\\tiny 2022}}  
                     & \textbf{Perception} / 3D Object Detection& \makecell[c]{Composite Detection Score (CDS)} &\multirow{5}{*}{81}  & \cmark \\
                    \cmidrule{4-5}
                     &   & 
                     &   \textbf{Prediction} / Motion Forecasting  & \makecell[c]{Average Brier Minimum FinalDisplacement\\Error (avgBrierMinFDE)}&    &  \cmark \\
                     \cmidrule{4-5}
                    &   & 
                    &   \textbf{Perception} / Stereo Depth Estimation &  Number of Bad Pixels &   & \cmark \\
        
                    \cmidrule{3-7}
        
                     &  & \multirow{5}{*}{\makecell[l]{\tiny CVPR\\\tiny 2021}} 
                     & \textbf{Perception} / Stereo Depth Estimation&Number of Bad Pixels &\multirow{5}{*}{368}  & \cmark \\
                    \cmidrule{4-5}
                     &   &  
                     & \textbf{Prediction} / Motion Forecasting & \makecell[c]{Brier Minimum Final Displacement Error\\(brier-MinFDE)} & &   \cmark \\
                     \cmidrule{4-5}
                    &   &
                    & \textbf{Perception} / Streaming 2D Detection & AP & &     \cmark \\

                \arrayrulecolor[gray]{0.9}\specialrule{3pt}{0pt}{-6pt} \arrayrulecolor{black}
                 \midrule 

        \rowcolor[gray]{0.9}
                        &   & \tiny{2023}  
                    & \textbf{Planning} / CARLA AD Challenge 2.0 & \ \ \ \ \ \ \ \ \ \ \makecell[c]{Driving Score, Route Completion,\\Infraction Penalty} \ \ \ \ \ \ \ \ \ \ \  & - & \cmark \\
        \arrayrulecolor[gray]{0.9}\specialrule{6pt}{0pt}{-6pt} \arrayrulecolor{black}
                    \cmidrule{3-7}
        \rowcolor[gray]{0.9}
                     &  & \makecell[l]{\tiny NeurIPS\\\tiny 2022} \ 
                     &\textbf{Planning} / CARLA AD Challenge 1.0 & \ \ \ \ \ \ \ \ \ \ \makecell[c]{Driving Score, Route Completion,\\Infraction Penalty} \ \ \ \ \ \ \ \ \ \ \  & 19 & \cmark \\
        \arrayrulecolor[gray]{0.9}\specialrule{6pt}{0pt}{-6pt} \arrayrulecolor{black}
                    \cmidrule{3-7}
        \rowcolor[gray]{0.9}
                     \multirow{-6}{*}{\makecell[l]{\tiny CARLA\\\tiny Autonomous\\\tiny Driving\\\tiny Challenge\\\tiny \citep{adchallengecarka, dosovitskiy2017carla}}}& \multirow{-6}{*}{\makecell[l]{\tiny CARLA\\\tiny Team, Intel}} & \makecell[l]{\tiny NeurIPS\\\tiny 2021} \ 
                     & \textbf{Planning} / CARLA AD Challenge 1.0 & \ \ \ \ \ \ \ \ \ \ \makecell[c]{Driving Score, Route Completion,\\Infraction Penalty} \ \ \ \ \ \ \ \ \ \ \  & - & \cmark \\
        
        \arrayrulecolor[gray]{0.9}\specialrule{3pt}{0pt}{-3pt} \arrayrulecolor{black}

                     \midrule 
        
                    \multirow{6}{*}{\makecell[l]{\tiny International\\\tiny Algorithm\\\tiny Case\\\tiny Competition\\\tiny\citep{adchallengepazhou}}}    & \multirow{6}{*}{\makecell[l]{\tiny Pazhou Lab}}  & \multirow{3.5}{*}{\makecell[l]{\tiny 2023}}  
                    &\textbf{Perception} / Monocular Depth Estimation & Accuracy, Efficiency Cost & - & \xmark \\
                    \cmidrule{4-5}
                     &   & 
                     &  \makecell[l]{\textbf{Perception} / Roadside Radar Calibration\\and Object Tracking} & \makecell[c]{Average Accuracy, Real-time Complexcity}&   -   & \xmark \\
        
                    \cmidrule{3-7}
        
                     &  & \multirow{2}{*}{\makecell[l]{\tiny 2022}}  
                     &\textbf{Perception} / Roadside 3D Perception & \makecell[c]{mAP, Similarity}&-  &  \xmark \\
                    \cmidrule{4-5}
                      &   &  
                      & \textbf{Perception} / OCR on Street-view Image & Precision, Recall &   - & \xmark \\

                    \arrayrulecolor[gray]{0.9}\specialrule{3pt}{0pt}{-6pt} \arrayrulecolor{black}
                    \midrule 

        \rowcolor[gray]{0.9}
                      &   & \makecell[l]{\tiny NeurIPS\\\tiny 2021} \ 
                    &  \textbf{Perception} / nuScenes Panoptic  &  Panoptic Quality (PQ) & 11  & \cmark \\
        
                    \arrayrulecolor[gray]{0.9}\specialrule{6pt}{0pt}{-6pt} \arrayrulecolor{black}\cmidrule{3-7}
        \rowcolor[gray]{0.9}
                     &  &  
                     & \textbf{Perception} / nuScenes Detection&nuScenes Detection Score (NDS)&  &  \cmark \\
                    \arrayrulecolor[gray]{0.9}\specialrule{6pt}{0pt}{-6pt} \arrayrulecolor{black}\cmidrule{4-5}
        \rowcolor[gray]{0.9}
                      &   &  
                      &  \textbf{Perception} / nuScenes Tracking  &Average Multi Object Tracking Accuracy   &     & \cmark \\
                     \arrayrulecolor[gray]{0.9}\specialrule{6pt}{0pt}{-6pt} \arrayrulecolor{black}\cmidrule{4-5}
        \rowcolor[gray]{0.9}
                     &   &  
                     &  \textbf{Prediction} / nuScenes Prediction &  Minimum Average Displacement Error&  & \cmark \\
                     \arrayrulecolor[gray]{0.9}\specialrule{6pt}{0pt}{-6pt} \arrayrulecolor{black}\cmidrule{4-5}
        \rowcolor[gray]{0.9}
                      \multirow{-8}{*}{\makecell[l]{\tiny AI Driving\\\tiny Olympics\\\tiny \citep{Dataset_nuScenes}}}&  \multirow{-8}{*}{\makecell[l]{\tiny ETH Zurich,\\\tiny University of\\\tiny Montreal,\\\tiny Motional}} & \multirow{-5.5}{*}{\makecell[l]{\tiny ICRA\\\tiny 2021}} 
                     &  \textbf{Perception} / nuScenes LiDAR Segmentation &mIoU&   \multirow{-5.5}{*}{456}    & \cmark \\
        
                     \arrayrulecolor[gray]{0.9}\specialrule{3pt}{0pt}{-3pt} \arrayrulecolor{black}

                    \midrule 
        
                   \makecell[l]{\tiny Jittor\\\tiny Artificial\\\tiny Intelligence\\\tiny Challenge\\\tiny\citep{adchallengejitu}}    & \makecell[l]{\tiny NSFC}  & \multirow{1}{*}{\makecell[l]{\tiny 2021}}  
                   & \textbf{Perception} /  Traffic Sign Detection  & Average Detection Accuracy & 37 & \xmark \\
        
                   \arrayrulecolor[gray]{0.9}\specialrule{3pt}{0pt}{-6pt} \arrayrulecolor{black}
                 \midrule 
        \rowcolor[gray]{0.9}
                    \makecell[l]{\tiny KITTI Vision\\\tiny Benchmark\\\tiny Suite\\\tiny\citep{ Dataset_kitti, geiger2012we,
                    Fritsch2013ITSC, Menze2015CVPR}}    &  \makecell[l]{\tiny University of\\\tiny Tübingen} \ \ &  \tiny 2012  
                    & \makecell[l]{\textbf{Perception} / Stereo, Flow, Scene Flow, Depth,\\Odometry, Object, Tracking, Road, Semantics}  & - & 5,610 &  \cmark \\    
                    \arrayrulecolor[gray]{0.9}\specialrule{3pt}{0pt}{-3pt} \arrayrulecolor{black}

                    \bottomrule            
        \end{tabular}
    }
    \underfigtab
\end{table}

\Cref{tab:tasks} presents a comprehensive overview of the primary international and domestic challenges and rankings in the realm of autonomous driving in recent years. The aforementioned in \Cref{tab:tasks} highlights the significant advancements and setbacks faced by the industry on a global and national level. The testing facility and list offer researchers a fair comparison platform for model outcomes, enabling the timely publication of model details and corresponding code. On certain highly active lists, the top position is frequently observed to change every month. Holding a competition and maintaining a list necessitate the combined efforts of organizers and other parties, including the provision of data download channels, preparation of benchmark models, and maintenance of test servers. 

In recent years, the prevalence of large models has escalated difficulties in holding competitions and maintaining lists. One of the challenges includes the growing data volume, which poses difficulty in downloading data. However, the demand for computing resources to train large models presents difficulties for organisations and individuals without access to a significant number of computing resource. Therefore, it is crucial for future competitions and rankings to appeal to a diverse range of researchers. Achieving this goal requires both the sponsors and participants to collaborate on initiatives such as improving the availability of data and ensuring the baseline models are user-friendly.

\subsection{Introduction to Various Types of Datasets}\label{sec:Introduction_to_Various_Types_of_Datasets}

Below, this section provides a detailed introduction and summary of various types of autonomous driving datasets, including perception, mapping, prediction, path planning, and interdisciplinary datasets, categorized by different tasks.
\subsubsection{Perception Datasets}\label{sec:Perception_Datasets}

\begin{table}[t]
    \centering
    \caption{
        \textbf{Perception Datasets at a glance.} 
        The table is divided into two parts, with the upper half mainly on perception tasks and the lower half on hybrid tasks. 
        For \texttt{Data Diversity}, \textit{Scenes} indicates the number of video clips.
        For \texttt{Region}, ``AS'' stands for Asia, ``EU'' for Europe, ``NA'' for North America, ``Global'' for all continents, and ``Sim'' for simulation scenarios. 
        For \texttt{Annotation}, ``2D/3D BBox'' indicates 2D/3D bounding box, ``2D/3D Seg'' means 2D/3D segmentation, ``3D Occ'' means 3D occupancy. 
        \texttt{Impact} implies the impact of the dataset defined in the context. 
        ``$-$'' indicates that a field is inapplicable.
    }
    \label{tab:dataset_comparison}
    \resizebox{\columnwidth}{!}{
        \footnotesize
        \begin{tabular}{l|c|c|c|c|c|c|c|c|c}
            \toprule
            \multirow{2}{*}{\textbf{Dataset}}  & \multirow{2}{*}{\textbf{Year}} & \multicolumn{3}{c|}{\textbf{Data Diversity}} &  
            \multicolumn{3}{c|}{\textbf{Sensor}} & \multirow{2}{*}{\textbf{Annotation}}  & \multirow{2}{*}{\textbf{Impact}}\\
            \cline{3-8}    &     & Scenes & Hours& Region& Camera &   LiDAR    & Other  &   &\\
\arrayrulecolor[gray]{0.9}\specialrule{3pt}{0pt}{-6pt} \arrayrulecolor{black}
            \midrule
        \rowcolor[gray]{0.9}     KITTI~\citep{geiger2012we}                &  2012 & 50        & 6    &  EU   &    Front-view   & \cmark     & GPS \& IMU          & 2D BBox \& 3D BBox          &\textbf{High}\\
             Cityscapes~\citep{Dataset_cityscapes}       &  2016 & -         & -      & EU    & Front-view     & \xmark      &           & 2D\ Seg         & Mid\\
             Lost and Found~\citep{Dataset_lost_and_found} & 2016 & 112 & - & - &Front-view  & \xmark & &2D\ Seg& \\
             Mapillary~\citep{Dataset_mapillary} & 2016 & - & - & Global&Street-view & \xmark &  & 2D\ Seg& Mid\\
             DDD17~\citep{Dataet_ddd17} & 2017 & 36 & 12 &EU & Front-view & \xmark & GPS\ \&\ CAN-bus\ \&\ Event Camera &-& \\     
             Apolloscape~\citep{Dataset_apolloscape} & 2016 & 103 & 2.5 & AS & Front-view & \xmark & GPS\ \&\ IMU   & 3D BBox\ \&\ 2D Seg & Mid \\
             BDD-X~\citep{Dataset_bdd100k-x} & 2018 & 6984 & 77 & NA &Front-view  & \xmark &  & Language &  \\
             HDD~\citep{Dataset_hdd}  & 2018 & - & 104  & NA & Front-view  & \cmark & GPS\ \&\ IMU\ \&\ CAN-bus& 2D\ BBox & Mid \\
             IDD~\citep{Dataset_idd}  & 2018 & 182 & -  & AS &Front-view &\xmark & &2D\ Seg &  \\
             SemanticKITTI~\citep{Dataset_semantickitti} & 2019 & 50 & 6 & EU  &\xmark & \cmark & & 3D Seg &  \\
             Woodscape~\citep{Dataset_Woodscape}  & 2019 & - & - & Global  & 360°  & \cmark & GPS\ \&\ IMU\ \&\ CAN-bus & 3D BBox\ \&\ 2D\ Seg & Mid \\
             DrivingStereo~\citep{Dataset_drivingstereo} & 2019 & 42 &- & AS & Front-view &\cmark  & &- &\\
             Brno-Urban~\citep{Dataset_Brno-Urban} & 2019 & 67 & 10 &EU & Front-view & \cmark & GPS\ \&\  IMU\ \&\ Infrared Camera& -&\\
             A*3D~\citep{Dataset_A*3D} & 2019 &- &55 &AS &Front-view &\cmark & & 3D BBox &Mid\\
             Talk2Car~\citep{Dataset_talk2car} & 2019 & 850 & 283.3 &NA & Front-view & \cmark &  & Language\ \&\ 3D BBox &  \\
             Talk2Nav~\citep{Dataset_talk2nav} & 2019 &10714 &- &Sim& 360°& \xmark
        &  & Language&\\
             PIE~\citep{Dataset_pie} & 2019 &- &6 &NA &Front-view &\xmark
         & & 2D\ BBox&\\
             UrbanLoco~\citep{Dataset_urbanloco} & 2019 &13 &-  &AS\ \&\ NA&360°&\cmark& IMU& -&\\
             TITAN~\citep{Dataset_titan} & 2019 &700 & -& AS&  Front-view&\xmark &  &2D\ BBox &\\
             H3D~\citep{Dataset_h3d} & 2019 & 160 & 0.77 & NA & Front-view & \cmark & GPS\ \&\ IMU & -&\\
        
             A2D2~\citep{Dataset_a2d2} & 2020 & - & 5.6 & EU & 360° & \cmark & GPS \& IMU \& CAN-bus & 3D BBox\ \&\ 2D\ Seg &  \\
             CARRADA~\citep{Dataset_carrada} & 2020 & 30&0.3 &NA & Front-view&\xmark &Radar & 3D BBox&\\
             DAWN~\citep{Dataset_dawn} & 2019 & -&- &Global & Front-view & \xmark& &2D\ BBox &\\
             4Seasons~\citep{Dataset_4seasons} & 2019 &- &- &-& Front-view&\xmark &GPS \& IMU &  -&\\
             UNDD~\citep{Dataset_undd} & 2019 & -&- &- & Front-view&\xmark & &2D\ Seg &\\
             SemanticPOSS~\citep{Dataset_semanticposs} & 2020 &- &- &AS &\xmark & \cmark&GPS \& IMU &3D Seg &\\
             Toronto-3D~\citep{Dataset_toronto-3d} & 2020 &4 &- &NA &\xmark & \cmark& &3D Seg &\\
             ROAD~\citep{Dataset_road} & 2021 & 22& -& EU& Front-view& \xmark& & 2D BBox \& Topology&\\
             Reasonable Crowd~\citep{Dataset_reasonable} & 2021 & - & -& Sim&Front-view &\xmark & & Language&\\
             METEOR~\citep{Dataset_meteor} & 2021 &1250 & 20.9&AS &Front-view& \xmark & GPS&Language &\\
             PandaSet~\citep{xiao2021pandaset} & 2021 & 179&- & NA& 360°& \cmark&GPS \& IMU & 3D BBox&\\
             MUAD~\citep{Dataset_muad} & 2022 &- & -&Sim &360°& \cmark & & 2D Seg \& 2D BBox&\\
             TAS-NIR~\citep{Dataset_tas-nir} & 2022 &- &- & -& Front-view&\xmark & Infrared Camera& 2D\ Seg&\\
             LiDAR-CS~\citep{Dataset_lidar-cs} & 2022 &6& -&Sim & \xmark&\cmark & &3D BBox &\\
             WildDash~\citep{Dataset_wilddash} & 2022 & -&- & -& Front-view&\xmark & &2D\ Seg&\\
             OpenScene~\citep{sima2023_occnet} & 2023 & 1000 & 5.5 & AS\ \&\ NA & 360° & \xmark &  & 3D Occ &  \\
             ZOD ~\citep{alibeigi2023zenseact} & 2023 & 1473 & 8.2 & EU & 360° & \cmark & GPS \& IMU \& CAN-bus &  3D BBox \& 2D Seg & Mid \\
\arrayrulecolor[gray]{0.9}\specialrule{3pt}{0pt}{-6pt} \arrayrulecolor{black}
            \midrule
        \rowcolor[gray]{0.9}     nuScenes~\citep{Dataset_nuScenes} & 2019 & 1000 & 5.5 & AS\ \&\ NA & 360° & \cmark & GPS\ \&\ CAN-bus\ \&\ Radar\ \&\ HDMap & 3D BBox\ \&\ 3D Seg &\textbf{High}  \\
        \rowcolor[gray]{0.9}     Argoverse V1~\citep{Dataset_argoversev1} & 2019 & 324k & 320 & NA  & 360° & \cmark & HDMap & 3D BBox\ \&\ 3D Seg & \textbf{High}  \\
        \rowcolor[gray]{0.9}     Waymo~\citep{Dataset_Waymo} & 2019 & 1000 & 6.4 & NA &360° & \cmark &  & 2D\ BBox\ \&\ 3D BBox & \textbf{High}\\
             KITTI-360~\citep{Dataset_kitti-360} & 2020  & 366 & 2.5 & EU & 360° & \cmark  & & 3D BBox \&\ 3D Seg &  Mid \\
             ONCE~\citep{Dataset_once} & 2021 & - & 144 & AS & 360° &  \cmark  &  & 3D BBox &   Mid\\
        \rowcolor[gray]{0.9}     nuPlan~\citep{Dataset_nuplan}  & 2021 & - & 120 & AS\ \&\ NA &  360° & \cmark &  & 3D BBox  & \textbf{High} \\
             Argoverse V2~\citep{Dataset_argoversev2} & 2022 & 1000 & 4 & NA &  360° & \cmark & HDMap   & 3D BBox  & Mid \\
            DriveLM~\citep{drivelm2023}  & 2023 & 1000 & 5.5 & AS\ \&\ NA &  360° & \xmark &  & Language  & Mid \\
            \bottomrule
        \end{tabular}
    }
    \underfigtab
\end{table}

\begin{figure}[t]
    \centering
    \includegraphics[width=0.8\linewidth]{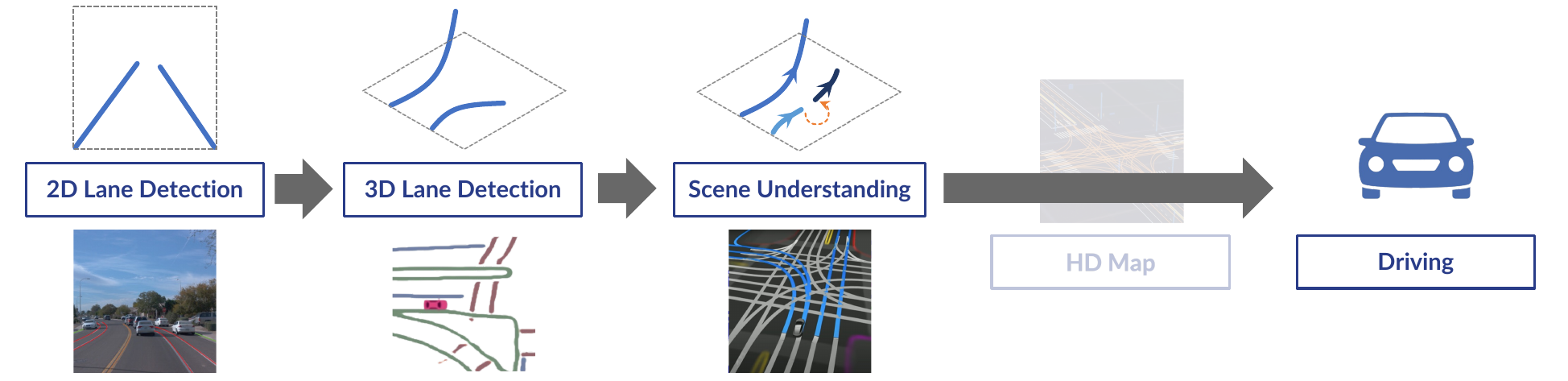}
    \caption{
        \textbf{Roadmap of Road Structure Understanding.}
        Instead of relying on HD maps, online mapping methods are more promising for autonomous driving in the future.
    }
    \label{fig:Roadmap_of_Road_Structure_Cognition}
    \underfigtab
\end{figure}


\Cref{tab:dataset_comparison} presents comprehensive information on the perception task datasets spanning from 2012 to 2023. It includes details on sensor configuration, tasks, evaluation indicators, and impact score, among others. The perception task evolves from traditional 2D object detection, to 3D object detection, to the newer occupancy grid detection~\citep{gao2023Sparse,huang2023leveraging,huang2023geometric,liu2023density,tong20233d,zeng2023distilling}.

Overall, it can be seen that the diversity of the data is greatly improved, which is mainly reflected in the duration of the datasets, the number of scenes and the geographical distribution of the data. Aside from the richness of data and sensors, algorithms guiding development schemes and technical paths influence the continuous iteration of data annotations.

\subsubsection{Mapping Datasets}\label{sec:Mapping_Datasets}

\begin{table}[t!] 
    \centering
    \caption{
        \textbf{Review of Mapping Datasets.}
        The representation format of ground truth is continuously evolving.
        Numbers in \textit{Frames} are the number of annotated frames over total frames respectively.
        \textit{Inst.} indicates whether lanes are annotated instance-wise (c.f. semantic-wise). \textit{Track.} implies whether a lane has a unique tracking ID.
    }
    \label{tab:dataset_mapping_comparison}
    \resizebox{\columnwidth}{!}{
        \footnotesize
        \begin{tabular}{l|c|c|c|c|c|c|c|c|c}
        \toprule
        \multirow{2}{*}{\textbf{Dataset}}  & \multirow{2}{*}{\textbf{Year}} &  \multicolumn{2}{c|}{\textbf{Data Diversity}} &  
        \multicolumn{2}{c|}{\textbf{Sensor}}   & \multicolumn{4}{c}{\textbf{Annotation}}\\
        \cline{3-10}    &     & Scenes & Frames & Camera &   LiDAR     & Type&   Space & Inst. &   Track. \\
        \midrule
        Caltech Lanes~\citep{aly2008real}&  2008& 4& 1224/1224& \multirow{11}{*}{Front-view Image} & \xmark &\multirow{11}{*}{Laneline}& PV&\cmark & \xmark \\
        VPG~\citep{lee2017vpgnet} &2017  & - &20K/20K& & \xmark && PV& \xmark & - \\
        TUsimple~\citep{tusimple2017} &2017& 6.4K&  6.4K/128K && \xmark  && PV  & \cmark& \xmark \\
        CULane~\citep{pan2018spatial} &2018 &-&133K/133K&  & \xmark&  & PV & \cmark& - \\
        ApolloScape~\citep{Dataset_apolloscape}& 2018 &235&115K/115K&   & \cmark & &PV&\xmark & \xmark \\
        LLAMAS~\citep{llamas2019}& 2019 & 14 &79K/100K& & \xmark &&PV &\cmark & \xmark \\
         3D Synthetic~\citep{guo2020gen}&  2020&- &10K/10K&  & \xmark& &PV &\cmark & - \\
         CurveLanes~\citep{xu2020curvelane}&  2020& - &150K/150K& & \xmark && PV& \cmark& - \\
        VIL-100~\citep{zhang2021vil}&2021  &100  &10K/10K& & \xmark &  &PV& \cmark& \xmark \\
         OpenLane-V1~\citep{chen2022persformer}&2022 &1K&200K/200K&  & \xmark  & &3D &  \cmark& \cmark \\
        ONCE-3DLane~\citep{Dataset_ONCE-3DLanes}& 2022 & -&211K/211K&  & \xmark &  &3D &  \cmark& - \\
        \midrule
        OpenLane-V2~\citep{wang2023openlanev2} & 2023 & 2K &72K/72K& Multi-view Image & \xmark & \makecell[c]{Lane Centerline,\\Lane Segment}& 3D &  \cmark& \cmark \\
        \bottomrule
        \end{tabular}
    }
    \underfigtab
\end{table}

\Cref{tab:dataset_mapping_comparison} outlines the fluctuating expression form of the ground truth of the dataset, specifically for the online mapping task. The progression of computer vision commences with image classification, detection, and segmentation, while mapping datasets originate from 2D image expressions like traffic light detection datasets. \Cref{fig:Roadmap_of_Road_Structure_Cognition} illustrates the representation and advancement of road structure cognition. With the expansion of mapping datasets, the data inferred by the model's predictive outcomes increasingly resembles the accurate and High-definition map(HD map). Currently, map perception results contain significantly less information than high-definition maps. Research is required to identify which characteristics of map perception results are most necessary for effective utilization by downstream tasks~\citep{lu2023translating}.

\subsubsection{Prediction and Planning Datasets}\label{sec:Predcition_and_Planning_Datasets}

\begin{table}[t]
    \centering
    \caption{
        \textbf{Review of Prediction and Planning Datasets.}
        The tasks are represented differently according to task formulations. 
    }
    \label{tab:planning}
    \resizebox{\columnwidth}{!}{
        \footnotesize
        \begin{tabular}{llllll}
        \toprule
          \textbf{Subtask}    & \textbf{Input} & \textbf{Output} & \textbf{Evaluation} & \textbf{Dataset}  & \textbf{Reference} \\
          
        \midrule
        
        \multirow{5}{*}{\makecell[l]{Motion\\ Prediction}} & \multirow{5}{*}{\makecell[l]{Surrounding Traffic\\States}}  & \multirow{5}{*}{\makecell[l]{Spatiotemporal\\Trajectories of\\Single/Multiple\\Vehicle(s)}} & \multirow{4}{*}{\makecell[l]{Displacement\\Error}} & Argoverse~\citep{Dataset_argoversev1} &~\citep{zhou2023qcnext},~\citep{feng2023macformer},~\citep{rowe2023fjmp}\\
             &   &  &  & nuScenes~\citep{Dataset_nuScenes} &~\citep{wu2022parallelnet},~\citep{zhang2022beverse},~\citep{park2023leveraging}  \\
             &   &  & & Waymo~\citep{Dataset_Waymo} &~\citep{shi2023mtr++},~\citep{sun2022m2i},~\citep{jia2023towards}\\
             &   &  & & Interaction~\citep{Dataset_interaction} &~\citep{jia2023hdgt},~\citep{vishnu2023improving},~\citep{jia2022multi}  \\
             &   &  & & MONA~\citep{gressenbuch2022mona} &  \\
             
        \midrule
        
        \multirow{4}{*}{\makecell[l]{Trajectory\\Planning}} &   \multirow{4}{*}{\makecell[l]{Motion States for Ego\\Vehicles, Scenario\\Cognition and\\Prediction}}    &   \multirow{4}{*}{\makecell[l]{Trajectories for\\Ego Vehicles}} & \multirow{4}{*}{\makecell[l]{Displacement Error,\\Safety, Compliance,\\Comfort}}  & 
        nuPlan~\citep{Dataset_nuplan} &~\citep{dauner2023parting},~\citep{phan2023driveirl},~\citep{xiimitation}  \\
        &   &  & & CARLA~\citep{dosovitskiy2017carla} &~\citep{zhang2021end},~\citep{chitta2021neat},~\citep{shao2023reasonnet}  \\
        &   &  & & MetaDrive~\citep{li2022metadrive} &~\citep{wang2023efficient},~\citep{wang2023autonomous},~\citep{liu2023guide}  \\
        
        &   &  & & Apollo~\citep{huang2018apolloscape} &~\citep{li2019aads},~\citep{piazzoni2020modeling},~\citep{bi2023application}  \\
        \midrule
        \multirow{5}{*}{\makecell[l]{Path\\Planning}}& \multirow{5}{*}{\makecell[l]{Maps for Road\\Network}} & \multirow{5}{*}{\makecell[l]{Routes Connecting\\to Nodes and Links}} &\multirow{5}{*}{\makecell[l]{Efficiency, Energy\\Conservation}}    & OpenStreetMap~\citep{haklay2008openstreetmap} &~\citep{hentschel2010autonomous},~\citep{luxen2011real},~\citep{de2016real}  \\
            &   &  &  & Transportation Networks~\citep{Transportation2023} &~\citep{guo2021gp3},~\citep{wang2023transportation},~\citep{patil2021convergence}   \\
            &   &  & & DTAlite~\citep{zhou2014dtalite} &~\citep{fu2023incremental},~\citep{xiong2018agbm},~\citep{tong2019open}\\
        
             &   &  & & PeMS~\citep{pems2017caltrans} &~\citep{zeng2021multi},~\citep{khan2017real},~\citep{wei2023evaluating}\\
             &  &  & & New York City Taxi Data~\citep{donovan2014new} &~\citep{ferreira2013visual},~\citep{zhu2018online},~\citep{tran2022adaptive}  \\
        \bottomrule
        \end{tabular}
    }
    \underfigtab
\end{table}

The traditional modular approach to regulatory technology divides the regulatory task into sub-tasks of varying dimensions. These sub-tasks are further categorised into three levels: road network, road, and vehicle. At the network level, traffic flow prediction and route planning can utilise static road topology data from the HD map and real-time road flow data from floating vehicles or intersections. The vehicle's driving route can then be planned from its starting point to its destination, generally with the aim of minimising travel time or driving distance. At road level, guided by the purpose of route guidance, the system predicts the driving behaviour and movement trajectory of surrounding vehicles in the next few seconds, based on perception information from a small range of road scenes near the self-driving car provided by sensors such as on-board cameras, LiDAR and radar. Afterward, it plans a self-driving car movement trajectory that is safe, efficient and comfortable. The vehicle level, in conjunction with the vehicle's kinematics and dynamics model, calculates the required accelerator and brake pedal input, as well as the steering wheel or steering wheel angle, to govern the vehicle's acceleration, deceleration, and steering. This allows the vehicle to follow the intended trajectory at the lowest possible control cost. Refer to \Cref{tab:planning} for the datasets related to the autonomous driving forecasting and planning task.

\subsubsection{Interdisciplinary Datasets}\label{sec:Interdisciplinary_Datasets}

\begin{table}[t!]
    \centering
    \caption{
        \textbf{Interdisciplinary Datasets and their representative works.}
        Dataset requirements vary across different fields.
    }
    \label{tab:Interdisciplinary_Datasets}
    \resizebox{\columnwidth}{!}{
        \footnotesize
        \begin{tabular}{llll}
            \toprule
              \textbf{Field}    & \textbf{Description} & \textbf{Dataset}  & \textbf{Reference} \\
              
            \midrule
            
            Robotics         & \makecell[l]{Replacing People for Multiple Operations\\and Services} & \makecell[l]
            {Roboturk~\citep{mandlekar2018roboturk},  PaLI-X~\citep{chen2023palix}, Robonet~\citep{dasari2019robonet}}&~\citep{driess2023palme},~\citep{kumra2017robotic},~\citep{pan2022iso}\\
            
            \midrule
            
            V2X              & \makecell[l]{Collaborative Sensing and Decision Making\\based on Wireless Communication} & \makecell[l]{Dair-V2X~\citep{yu2022dair}, V2X-Sim~\citep{li2022v2x}, OPV2V~\citep{xu2022opv2v}} &~\citep{zhang20233d},~\citep{ren2023interruption},~\citep{hu2022where2comm}\\
            
            \midrule

            UAV              & \makecell[l]{Aircraft with Multiple Sensors for Aerial\\Operations} & \makecell[l]{UAV123~\citep{mueller2016benchmark}, Blackbird~\citep{antonini2020blackbird}, UAVid~\citep{lyu2020uavid}}&~\citep{an2023learning},~\citep{cioffi2023hdvio},~\citep{li20222dsegformer}\\
            
            \midrule
            
            USV              & \makecell[l]{Operation and Control of the Unmanned\\Surface Vehicles} & \makecell[l]{MODS~\citep{bovcon2021mods}, MaSTr1325~\citep{bovcon2019mastr1325}, USVInland~\citep{cheng2021we}}&~\citep{han2023towards},~\citep{yao2021shorelinenet},~\citep{yin2023improved}\\
            
            \midrule
            
            Transportation        & \makecell[l]{Allocation and Coordination of Products,\\Raw Materials, and Services in the Supply\\Chain} & \makecell[l]{NEO Benchmark Datasets~\citep{neo2013}, LARa~\citep{niemann2020lara},  Divvy data~\citep{Divvy2015}}&~\citep{niazy2020solving},~\citep{kirchhof2021chances},~\citep{yang2020exploring} \\
            \bottomrule
        \end{tabular}
    }
    \underfigtab
\end{table}

The technical relationship between autonomous driving and other application areas deserves attention. It also highlights the positive impact of these datasets in various application fields on autonomous driving. \Cref{tab:Interdisciplinary_Datasets} presents pertinent datasets for Robotics, Vehicle-to-everything (V2X), Unmanned Aerial Vehicle (UAV), Unmanned Surface Vehicle (USV), Transportation, and other fields, respectively.

\section{Data Engine} \label{sec:Data_Engine}

\textbf{Efficiently constructing massive and high-quality data remains an open challenge.} Academia and industry implement distinct strategies to produce autonomous driving datasets, resulting in variations in data collection, quality control, annotation technology, \etc, depending on their respective platforms and technologies. In this chapter, we will compare data engine systems across different manufacturers to identify their commonalities and essential elements. At the same time, we will also examine each of these vital technologies to establish a robust basis for the following era of self-driving data collections.

\subsection{Commercial Platforms}\label{sec:Commercial_Platforms}

\begin{table}[t]
    \centering
    \caption{
        \textbf{Comparison of data engine systems.}
        A consensus on implementation among different entities has yet to be reached.
    }
    \label{tab:loop:business}
    \resizebox{\columnwidth}{!}{
        \footnotesize
        \begin{tabular}{lc|c|c|c|c|c|c}
            \toprule
             \textbf{Solution} & \textbf{Owner} & \textbf{Data Catelog} & \textbf{Data Retrieval} & \textbf{Auto-labeling} & \textbf{Model Training} & \textbf{Simulation} & \textbf{Open-source} \\
            \midrule
             \makecell[l]{Autonomous\\Driving Data\\Framework\\(ADDF)~\citep{addfamazon}} & \makecell[c]{
             \includegraphics[width=.06\textwidth]{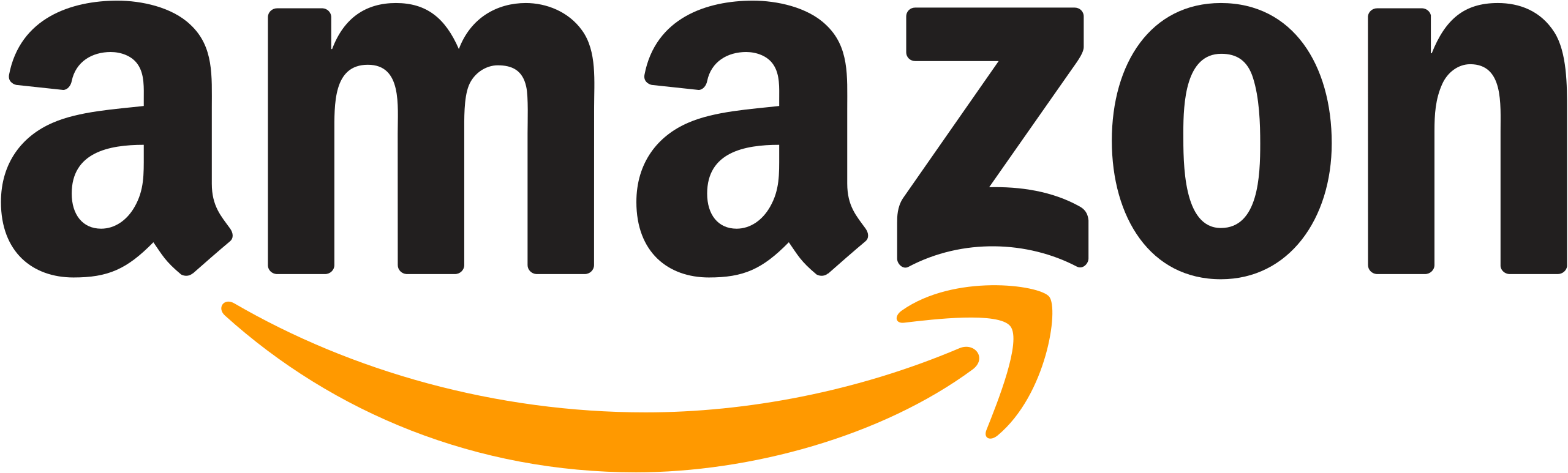}} & \makecell[l]{Scene Description\\via Scene Metadata\ \ \ \ \ \ \ \ }  & \makecell[l]{OpenSearch \ \ \ \ \ \ \ \ \ \ \ \ \ \  } & \makecell[l]{Object and Lane\\Detection via\\Open-source\\Models} & \cmark & \cmark & \cmark \\
             \midrule
            \makecell[l]{Full Self\\Driving\\(FSD)~\citep{fsdtesla}} & \makecell[c]{
            \includegraphics[width=.05\textwidth]{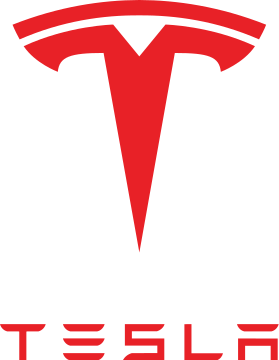}} & \multicolumn{2}{l|}{\makecell[l]{Misprediction Identification, Label Correction,\\and Selection on Most Valuable Examples}}  & \makecell[l]{Static Scene\\Annotation via\\Multi-trip\\ Reconstruction} & \makecell[l]{Dojo Supercomputer \ \ \ \ \ } & \makecell[l]{Scene Generation\\in Minutes} & \xmark \\
            \midrule
            \makecell[l]{MagLev~\citep{maglevnvidia}} & \makecell[c]{
            \includegraphics[width=.07\textwidth]{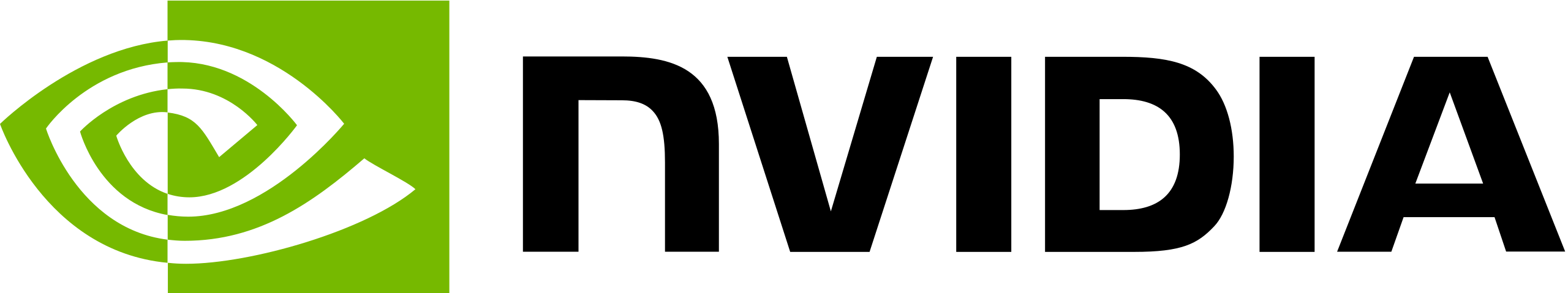}} & \makecell[l]{Generating Dataset\\via Searching, Collection,\\Labeling, and Export}   & \makecell[l]{Elastic Search and\\Categorization via\\Active Learning \ \ \ \ \ \ \ \ \ } & - & \makecell[l]{Multi-node Training\\and Parallel Evaluation} & \makecell[l]{DRIVE Sim \ \ \ \ \ \ \ \ } & \xmark \\
            \midrule
             \makecell[l]{OpenTrek~\citep{opentrek}} & \makecell[c]{
             \includegraphics[width=.07\textwidth]{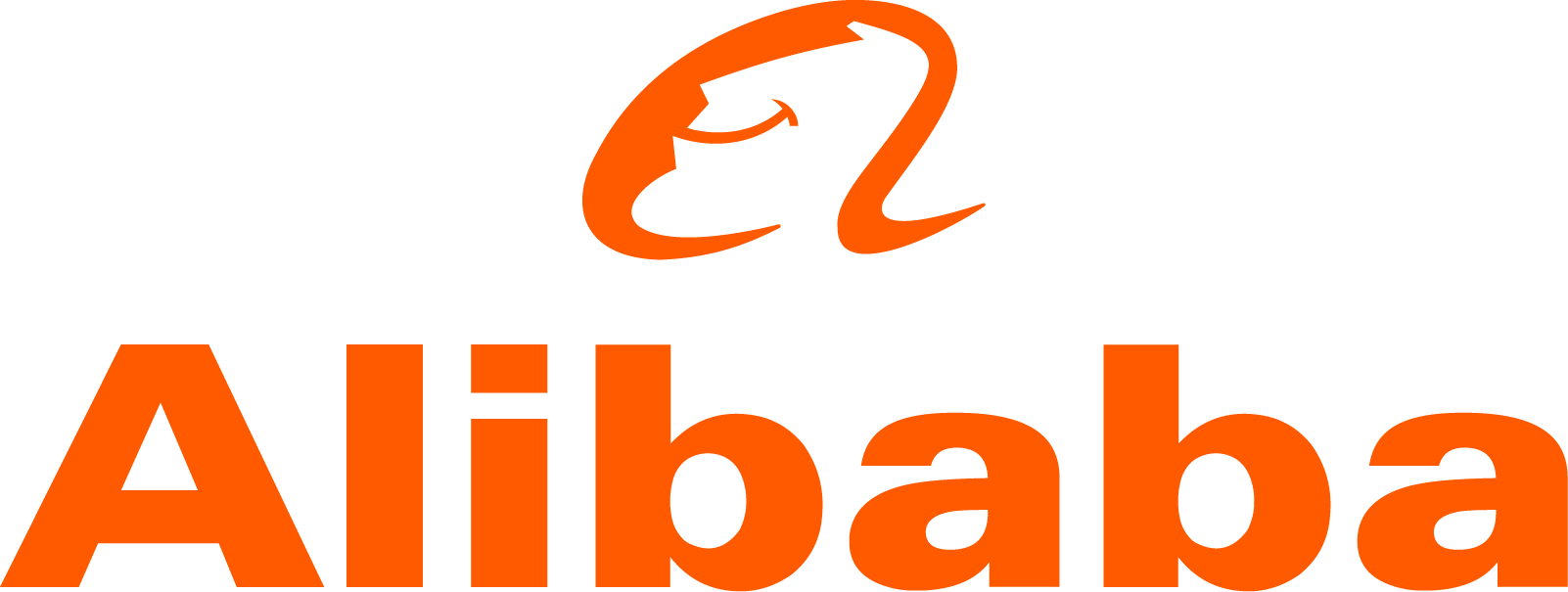}} & - & \makecell[l]{Multi-modal Retrieval\\based on Semantics,\\Images, Labels,\\Similarity, \etc} & - & \cmark & \cmark & \xmark \\
            \bottomrule
        \end{tabular}
    }
    \underfigtab
\end{table}

Data plays a significant role in the development cycle of autonomous vehicles. As demonstrated in \Cref{tab:loop:business}, relevant vendors establish their own closed-loop system of data algorithms to maximize their worth.
The data engine system usually encompasses data acquisition, preprocessing, storage, visualization, retrieval, labeling, usage, and more, with its aim being to effectively utilize vast amounts of data to enhance the dependability of the autonomous driving system.

\subsection{Data Labeling Tools}\label{sec:Data_Labeling_Tools}

\begin{table}[t!]
    \centering
    \caption{
        \textbf{Comparison of annotation toolkits.}
        \textit{2D/3D Seg} means the 2D or 3D segmentation label. 
        \textit{2D/3D BBox} is for 2D or 3D bounding box.
        \textit{2D/3D BBox-c} indicates 2D or 3D bounding box label task for continuous frame data.
        ``MA'' represents manual annotation mode, while ``SA'' represents semi-automatic annotation mode. Under \texttt{Point Cloud} and \texttt{Image}, \textit{Price} means the price for annotating a single instance. 
        \texttt{Toolkit Price} refers to the cost of using annotation tools.
        ``-'' implies that related information is inaccessible to the public.
        ``\xmark'' denotes that the target tool does not support this feature. 
    }
    \label{tab:label_tool}
    \resizebox{\columnwidth}{!}{
        \footnotesize
        \begin{tabular}{cc|cc|cc|c|cc|cc|c|c|c}
    
            \toprule
             \multirow{2}{*}{\textbf{Company}}     & \multirow{2}{*}{\textbf{Owner}}  & \multicolumn{5}{c|}{\textbf{Point Cloud}}  & \multicolumn{5}{c|}{\textbf{Image}} &\multirow{2}{*}{\textbf{Toolkit Price}}&\multirow{2}{*}{\textbf{\makecell[c]{Price of\\Sample Task}}}\\
             \cline{3-12} &    & \textbf{3D Seg}&\textbf{Price}  & \textbf{3D BBox} &\textbf{Price} &\textbf{3D BBox-c}   &\textbf{2D Seg} &\textbf{Price} &\textbf{2D\ BBox}  &\textbf{Price} &\textbf{2D BBox-c}&\\
             
            \midrule
            
            \makecell[c]{Sagemaker Ground\\Truth~\citep{sagemakeramazon}}& \makecell[c]{
            \includegraphics[width=.07\textwidth]{figures/logo/amazon.png}} &MA&-&\textbf{SA}&-&\textbf{SA}&\textbf{SA}&5.360&\textbf{SA}&0.255&\textbf{SA}&
            \makecell[l]{
            \textbullet \ \ For Each Point Cloud Frame:\\
             \ \ \ \ 3.00 USD for 1st Frame \\ 
             \ \ \ \ 1.50 USD for 2nd Frame Onwards\\
            \textbullet \ \ For Each Instance in Images:\\
             \ \ \ \ 0.08 USD less than 50K\\
            \ \ \ \ 0.04 USD between 50K and 1M	\\
            \ \ \ \ 0.02 USD more than 1M }&624K RMB \\
            \midrule
    
            \rule{0pt}{12pt}
            Baidu Cloud~\citep{zhinengyunbaidu} & \makecell[c]{
            \includegraphics[width=.08\textwidth]{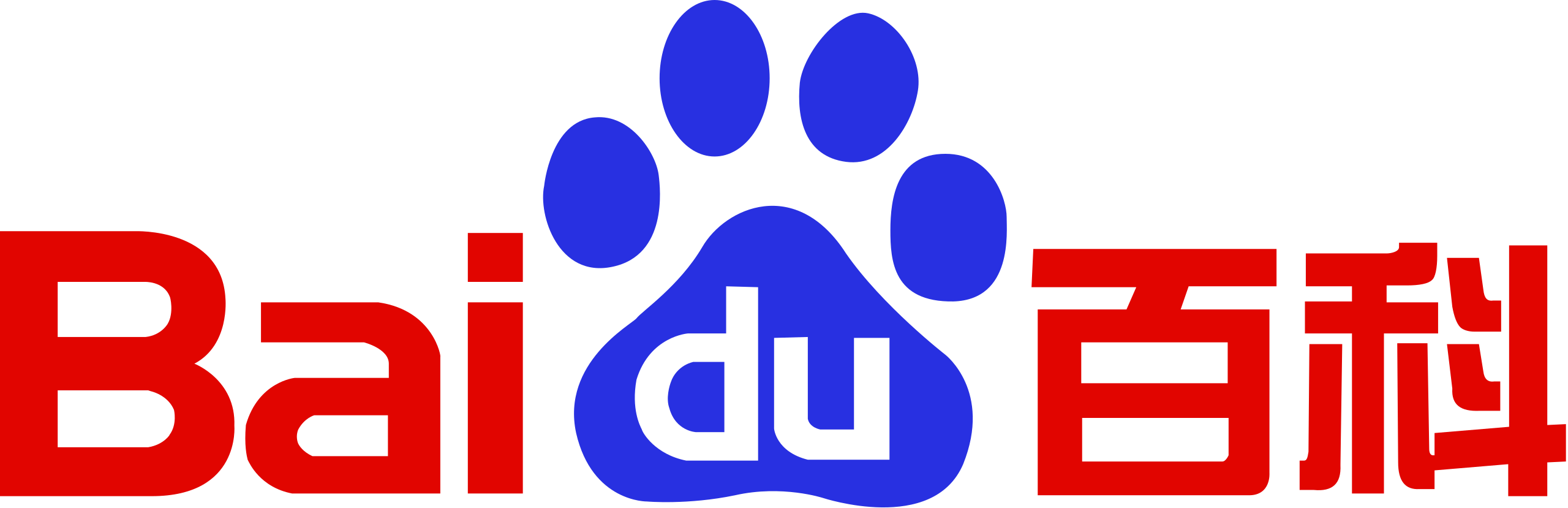}} &MA&0.3&MA&0.2-0.25&MA&MA&0.2&MA&0.06-0.1&MA &\makecell[c]{-}&93K RMB \\

              \midrule
    
            VOTT~\citep{vottmicrosoft} & \makecell[c]{
            \includegraphics[width=.08\textwidth]{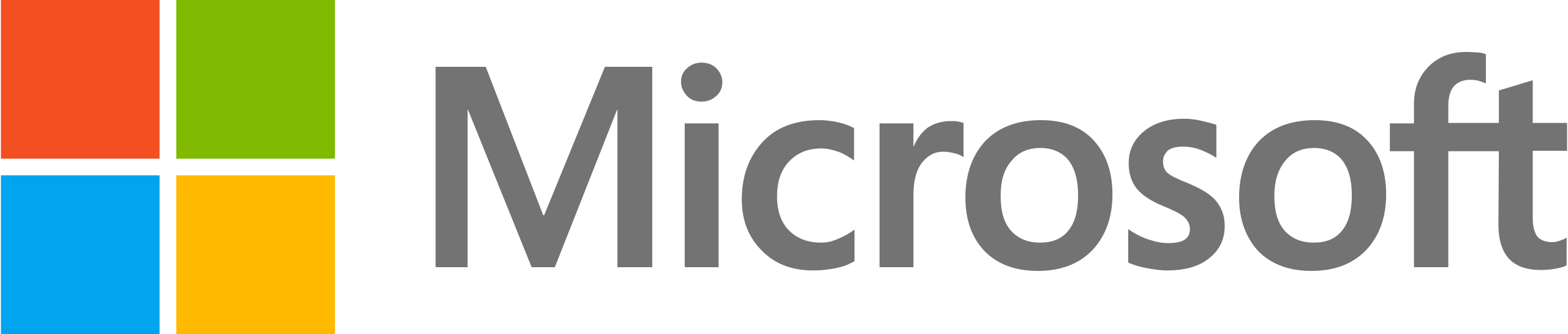}}  &\xmark&-&\xmark&-&\xmark&MA&-&\textbf{SA}&- &\textbf{SA}& \makecell[l]{Free - Open-source \ \ \ \ \ \ \ \ \ \ \ \ \ \ \ \ \ \ \ \ \ \ }&-\\
              \midrule
    
            \rule{0pt}{14pt}
            Data Pro~\citep{datatang} & \makecell[c]{
            \includegraphics[width=.08\textwidth]{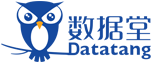}}  &\textbf{SA}&0.25&\textbf{SA}&0.3&\textbf{SA}&\textbf{SA}&0.25&\textbf{SA}&0.07 &SA& \makecell[c]{-}&90K RMB\\
              \midrule

            \makecell[c]{Multi-sensor\\Labeling Platform~\citep{segmentsai}} & \makecell[c]{
            \includegraphics[width=.1\textwidth]{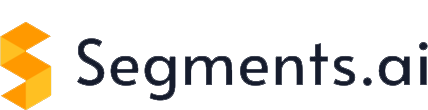}}  &\textbf{SA} & -& \textbf{SA}&- & \textbf{SA}&\textbf{SA} &- & \textbf{SA}& -& \textbf{SA}& - & -\\
              \midrule
    
            Data Engine~\citep{dataenginescale} & \makecell[c]{
            \includegraphics[width=.07\textwidth]{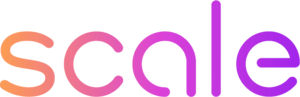}}  &\textbf{SA}&-&\textbf{SA}&-&\textbf{SA}&\textbf{SA}&18.2&\textbf{SA}&5.8 &\textbf{SA}&
            \makecell[l]{
             \textbullet \ \ Fixed Cost per Task: \\
            \ \ \ \ Based on the Setup of the Task\\
            \ \ \ \ at Task Creation\\
             \textbullet \ \ Variable Cost per Task: \\
            \ \ \ \ Depend on the Labelers' Response\\
            \ \ \ \ during the Task
            }&-\\
            \bottomrule
        \end{tabular}
    }
    \underfigtab
\end{table}

\Cref{tab:label_tool} compares multiple automatic annotation tools in terms of their technology capabilities, pricings, and functions. The tools discussed in this comparison include Amazon, Baidu, Microsoft, Datatang, Segment.ai and Scale AI. These annotation tools offer various functions that help automate and streamline the data annotation process and they support a range of annotation tasks, including image classification, object detection, and semantic segmentation. Some annotation tools incorporate semi-automatic annotation algorithms, including the 3D target annotation task for point cloud data. Amazon Sagemaker Ground Truth, Datatang Data Pro, and Scale AI's Data Engine annotation platform allow annotators to annotate only a portion of the data or provide prompts to the algorithm through mouse clicks to generate a complete set of high-quality 3D boundary box labels expediently. This process results in cost savings and improved efficiency. The labeling costs applicable are presented in \Cref{tab:label_tool}. Multi-sensor Labeling Platform, Sagemaker Ground Truth and Data Engine are charged separately, while Visual Object Tagging Tool (VOTT) is an open source tool.

\subsection{Case Study on Data Engine} \label{sec:Data_Engine_Tutorial}

\begin{figure}[t!]
    \centering
    \includegraphics[width=0.9\linewidth]{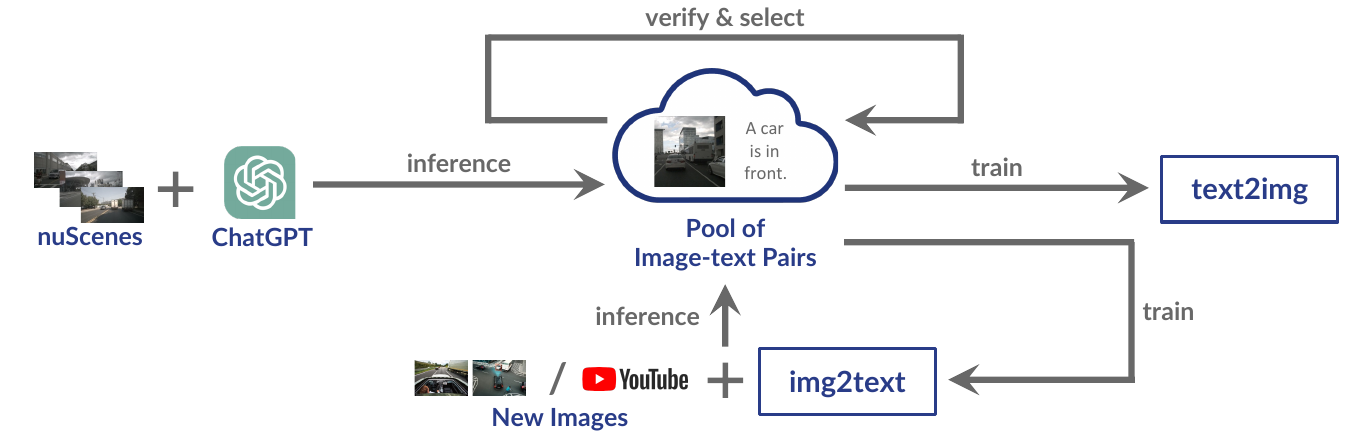}
    \caption{
        \textbf{Case study on Data Engine.}
        LLMs are integrated into the pipeline to generate high-quality image data conditioned on scene description.
    }
    \label{fig:applicationloop_v2}
    \underfigtab
\end{figure}

\begin{figure}[t]
    \centering
    \includegraphics[width=0.9\linewidth]{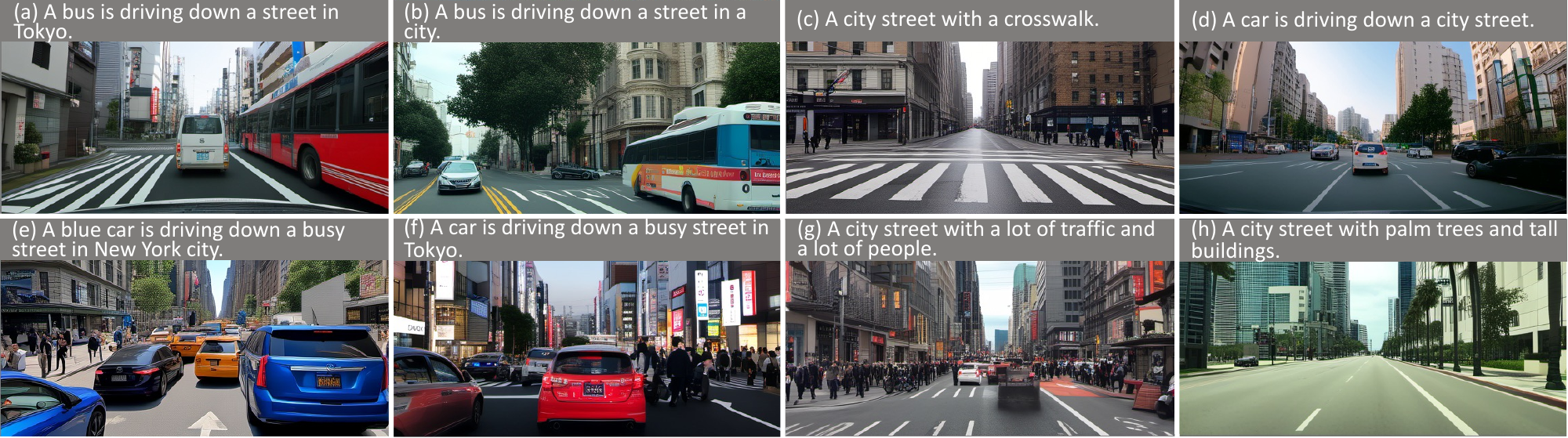}
    \caption{
        \textbf{Examples of generated images.}
        Based on the proposed pipeline, generated images are able to match the provided scene description.
    }
    \label{fig:application}
    \underfigtab
\end{figure}

In this chapter, we present a case study to demonstrate the solutions discussed herein. The overall progression of this particular case is depicted in \Cref{fig:applicationloop_v2}. A significant hurdle in this endeavour is the lack of integration between language processing and autonomous driving.
The objective of this study was to create an AIGC model that generates images relating to autonomous driving based on textual content.
As a result, current autonomous driving datasets generally lack language labels. To address this issue, the researchers employed the ChatGPT language model to automatically annotate images in the nuScenes dataset. This generated corresponding descriptive text labels. As ChatGPT was not originally developed for autonomous driving, the researchers utilized various templates to generate textual descriptions for an individual image. These were then thoroughly scrutinized and chosen several times in order to form a workable pair of data labels. After preparing the dataset, researchers utilise the annotated data to train the text-based generative model, text2image. However, the limited quantity of images in the nuScenes dataset and coverage of only a few scenes prevents the text2image model from generating pictures of certain scenes based on the text, such as dark and rainy nights.
To achieve this goal, the researchers incorporated external driving datasets, including extensive driving images from YouTube, to refine the existing img2text model into a model specifically for annotating images in the context of autonomous driving. Simultaneously, new images of autonomous driving were taken and extrapolated using the img2text model. From many image-text pairs, datasets that meet the criteria have been identified and reused to refine the model. As depicted in the image-to-text pool illustrated in \Cref{fig:application}, the manual intervention and data generation cycles alternate to steadily enhance the data quality in the pool.

\section{Towards Next-generation Autonomous Driving Datasets} \label{sec:Next-generation_Autonomous_Driving_Dataset}

\subsection{Inspirations from Foundation Models}

\begin{table}[t]
    \centering
    \caption{
        \textbf{Comparison of data and model parameters in the development of large models.}
        \texttt{\# Parameter} denotes the number of parameters in the neural network. 
        Under \texttt{\# Parameter}, ``B'' denotes billion, ``V'' means vision models, ``VL'' indicates vision-language models, and ``Gen'' represents generative models.
    }
    \label{tab:data_alg_solution}
    \resizebox{\columnwidth}{!}{
        \footnotesize
        \begin{tabular}{ccccccc}
            \toprule
            Field & Method & \makecell[c]{Organization/Date} & Data Scale & \# Parameter & Open-source & Feature \\
            \midrule
            \multirow{2.2}{*}{NLP} & GPT-4~\citep{openai2023gpt4} & OpenAI/2023.3 & \makecell[c]{13T Tokens} & 1800B (est.) & \xmark & Most Powerful LLM \\
            \cline{2-7}
            \rule{0pt}{10pt}
            & LLaMA 2~\citep{touvron2023llama} & Meta/2023.7 & 2T Tokens & 7-70B & \cmark & Open-sourced and full-stack LLM \\
            
            \midrule
            
            \multirow{3.5}{*}{General Vision} & ViT-22B~\citep{dehghani2023scaling} & Google/2023.2 & 4B Images & 22B & \xmark & \makecell[c]{Largest Vison-only Model\\ (Single-modal)} \\
            \cline{2-7}
            \rule{0pt}{17pt}
            & \makecell[c]{BLIP-2~\citep{li2023blip}\\ (Multi-modal)} & Salesforce/2023.1 & 129M Images & 12.1B & \cmark & \makecell[c]{LLM + Vision $\rightarrow$ VLM} \\
           
            \midrule
            
            \multirow{4.5}{*}{Robotics} & PaLM-E~\citep{driess2023palme} & Google/2023.3 & Unknown & \makecell[c]{562B}& \xmark & LLM + Robotics \\
            \cline{2-7}
            \rule{0pt}{22pt}
            & RT-2~\citep{brohan2023rt} & DeepMind/2023.7 & \makecell[c]{1B Image-Text\\Pairs, 13 Robots,\\17 Months} & 5/12/55B & \xmark & LLM + Robotics \\
            
            \midrule
            
            \multirow{1}{*}{Medical} & Med-PaLM 2~\citep{singhal2023towards} & Google/2023.4 & 1M Samples & 562B & \xmark & LLM + Med \\
    
            \midrule
            
           \multirow{4}{*}{Autonomous Driving} & 
           \makecell[c]{World Model\\Demo~\citep{teslaworldmodel}} & Tesla/2023.9.30 & Unknown & 1.1B (est.) & \xmark & \makecell[c]{World Model in Autonomous Driving} \\
            \cline{2-7}
            \rule{0pt}{22pt}
           & 
           DriveAGI~\citep{driveagi} & \makecell[c]{OpenDriveLab\\/Scheduled for 2024} & \makecell[c]{4M Images,\\LiDAR, \etc\\+ 32M Images} & \makecell[c]{1.2B (V/VL)\\+ 2.6B (Gen)} & \cmark & \makecell[c]{Reasoning-oriented, Multi-modal,\\Closed-loop: Perception + Generation} \\
            
            \bottomrule
        \end{tabular}
    }
    \underfigtab
\end{table}

\textbf{General large model vs autonomous driving large model.} \Cref{tab:data_alg_solution} displays the current impressive advancements in natural language processing, computer vision and other fields achieved by the large model, thereby initiating a new wave of research in artificial intelligence. It is evident that the large model of autonomous driving has considerable room for development in regard to data volume and model scale. Drawing from the use of large models in other sectors for reference, the new generation of datasets should strive to enhance data volume to be in line with those in other fields, to enable the use of large models in autonomous driving. It should be noted that in many fields, the implementation of large models introduces natural language, which is often biased towards general large models. As an autonomous driving system receives sensor data as input and produces vehicle path planning as output, the integration of natural language is not immediately evident. Additionally, the necessity of incorporating a universal grand model for autonomous driving is worth discussing.

\textbf{Data volume vs algorithm performance.} Despite the fact that infinite data can bring neural networks accuracy near 100\%, the expense of acquiring such a large quantity is also immeasurable. Previous research suggests that as the scale of data increases\citep{goyal2019scaling,brown2020language,dosovitskiy2020image,singh2023effectiveness,10132530}, model performance can significantly improve. However, once the data volume reaches a certain level, the growth in model performance tends to plateau. Therefore, the widespread implementation of autonomous driving technology mandates that the model performs accurately in rare situations to evade perilous or operational mishaps. Hence, in the context of autonomous driving, there is no need to indiscriminately augment the quantity of data. For the majority of circumstances encountered on the road, an immense amount of data is not essential to support the dependable functioning of automated driving. The long-tail scenario requires more attention due to infrequent traffic scenarios, such as crashes, which result in a lack of relevant data. Consequently, this lack of data significantly impacts the performance of the automatic driving system. Generally, algorithm performance is more reliant on scene richness whilst ensuring an adequate quantity of data.

\subsection{Essentials for Constructing the Next-generation Dataset}\label{sec:Next_Generation}

The initial and subsequent generations of datasets for autonomous driving are no longer capable of fulfilling the development requirements for autonomous driving systems. Consequently, the creation of a new era of dataset is imminently required. This chapter outlines the development objectives of the next generation of autonomous driving datasets.

\begin{enumerate}
    \setlength{\itemsep}{5pt}
    \setlength{\parsep}{0pt}
    \setlength{\parskip}{0pt}
    \item The development of large model has become an essential characteristic of these new datasets given the extensive advancement of larger models.
    \item The modular design of the autonomous driving system during the landing phase faced high iteration costs, performance limitations and other challenges. The end-to-end autonomous driving architecture is becoming increasingly popular in the industry~\citep{jia2023driveadapter,chen2022level}.
    \item The aspects of multi-modal sensors, high-quality labelling, and model logic reasoning ability require careful consideration~\citep{yang2023survey}.
\end{enumerate}

\section{Conclusion} \label{sec:Conclusion}

This survey provides a detailed review of the current state and challenges of autonomous driving datasets. 
In this survey, we systematically summarize datasets utilized in the development of autonomous driving, and demonstrate the importance of challenges and leaderboards for the research community. 
Furthermore, we provide a comprehensive overview of the data engine system in the industry domain, and analyze the role and functionality of major components in the pipeline. 
With a focus on data engine systems and trending foundation models, we present our vision and plan for the next generation of autonomous driving datasets. 
Given the rapid development,
we hope this survey would promote the advancement of autonomous driving in both academia and industry.

\section*{Author Contributions}
\textbf{H. Li and Y. Qiao} led the project and provided overall mentorship.\\
\textbf{H. Li} wrote the introduction section.\\
\textbf{Y. Li} was responsible for drafting the entire article (Sections 2 to 4) and figures.\\
\textbf{H. Wang } completed the challenge and data engine system section.\\
\textbf{J. Zeng} was responsible for revising and formatting the entire survey.\\
\textbf{H. Xu} was responsible for the revision of the manuscript. \\
\textbf{P. Cai} wrote content on law and interdisciplinary datasets.\\
\textbf{D. Lin, L. Chen, J. Yan, F. Xu, and L. Xiong} advised from their expertise. \\
\textbf{J. Wang and C. Xu} made modifications to the overall structure.\\
\textbf{F. Zhu, T. Wang, F. Xia, B. Mu, and Z. Peng} provided suggestions for the next-generation dataset.

\section*{Acknowledgments}
We thank members from OpenDriveLab for the discussion. Especially, we appreciate Jiazhi Yang, Jiahao Wang,  Chonghao Sima, Tianyu Li, Bangjun Wang, Chengen Xie, Xiaosong Jia, Penghao Wu for providing materials in this work. We also express our thanks to colleagues from the industry, including Qian Zhang, Xiangyu Zhang, Chengxiao Sun, Haofen Wang, \etc, for giving their suggestions to the survey. We would like to thank John Lambert, Klemens Esterle, Daniel Bogdoll from the public community for their advice to improve the manuscript.

\section*{Changelog}
\begin{itemize}
    \item V2 (Jan 23 2024): 
    Revised information in some Tables received from the community feedback; manuscript template altered by credit of the Gemini report from Google DeepMind.
    \medskip
    \item V1 (Dec 6 2023): 
    Preliminary draft of the English version available on arXiv.
\end{itemize}


\bibliography{sample}  %
\bibliographystyle{unsrt}

\end{document}